\pdfoutput=1
\documentclass[sigconf]{acmart}

\AtBeginDocument{%
  }

\copyrightyear{2024}
\acmYear{2024}
\setcopyright{acmlicensed}
\acmConference[IWSPA '24] {Proceedings of the 10th ACM International Workshop on Security and Privacy Analytics}{June 21, 2024}{Porto, Portugal.}
\acmBooktitle{Proceedings of the 10th ACM International Workshop on Security and Privacy Analytics (IWSPA '24), June 21, 2024, Porto, Portugal}
\acmISBN{979-8-4007-0556-4/24/06}
\acmDOI{10.1145/3643651.3659896}

\usepackage{algorithm}
\usepackage{algpseudocode}
\usepackage{amsfonts}
\usepackage{amsthm}
\usepackage{subcaption}
\usepackage{multirow}
\usepackage{bm}
\usepackage{adjustbox} 
\usepackage{dirtytalk}

\newtheorem{definition}{Definition}

\newtheorem{theorem}{Theorem}

\begin{document}
\title[1-Diffractor: Efficient and Utility-Preserving Text Obfuscation]{1-Diffractor: Efficient and Utility-Preserving Text Obfuscation Leveraging Word-Level Metric Differential Privacy}
\author{Stephen Meisenbacher}
\affiliation{
  \institution{Technical University of Munich \\ School of Computation, Information and Technology}
  \city{Garching}
  \country{Germany}
}
\email{stephen.meisenbacher@tum.de}

\author{Maulik Chevli}
\affiliation{
  \institution{Technical University of Munich \\ School of Computation, Information and Technology}
  \city{Garching}
  \country{Germany}
}
\email{maulikk.chevli@tum.de}

\author{Florian Matthes}
\affiliation{
  \institution{Technical University of Munich \\ School of Computation, Information and Technology}
  \city{Garching}
  \country{Germany}
}
\email{matthes@tum.de}

\renewcommand{\shortauthors}{Stephen Meisenbacher, Maulik Chevli, \& Florian Matthes}
\begin{abstract}
The study of privacy-preserving Natural Language Processing (NLP) has gained rising attention in recent years. One promising avenue studies the integration of Differential Privacy in NLP, which has brought about innovative methods in a variety of application settings. Of particular note are \textit{word-level Metric Local Differential Privacy (MLDP)} mechanisms, which work to obfuscate potentially sensitive input text by performing word-by-word \textit{perturbations}. Although these methods have shown promising results in empirical tests, there are two major drawbacks: (1) the inevitable loss of utility due to addition of noise, and (2) the computational expensiveness of running these mechanisms on high-dimensional word embeddings. In this work, we aim to address these challenges by proposing \textsc{1-Diffractor}, a new mechanism that boasts high speedups in comparison to previous mechanisms, while still demonstrating strong utility- and privacy-preserving capabilities. We evaluate \textsc{1-Diffractor} for utility on several NLP tasks, for theoretical and task-based privacy, and for efficiency in terms of speed and memory. \textsc{1-Diffractor} shows significant improvements in efficiency, while still maintaining competitive utility and privacy scores across all conducted comparative tests against previous MLDP mechanisms. Our code is made available at: \url{https://github.com/sjmeis/Diffractor}.
\end{abstract}



\maketitle

\section{Introduction}
\begin{figure*}[htbp]
    \centering
    \includegraphics[scale=0.45]{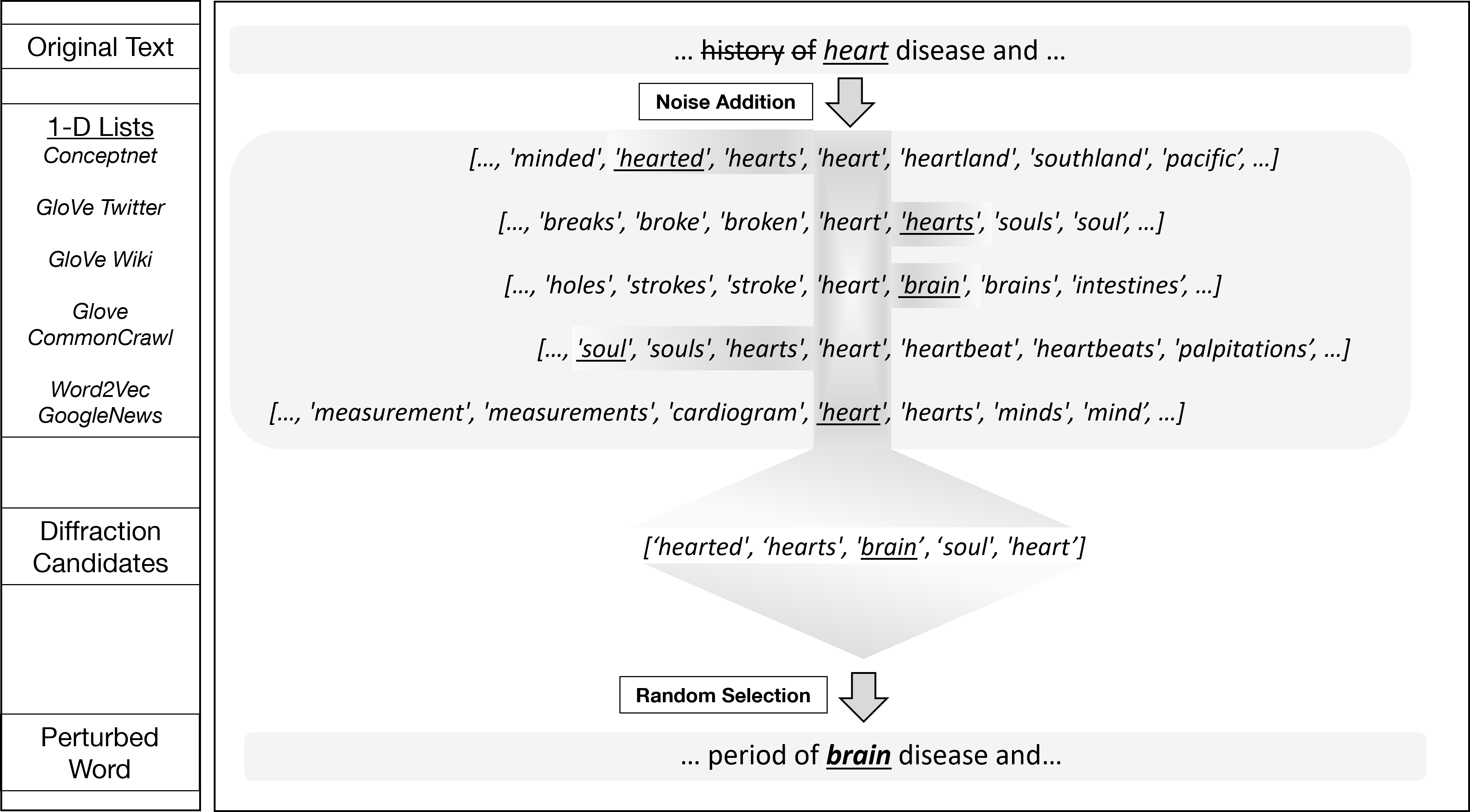}
    \caption{An Overview of \textsc{1-Diffractor}. Input text is perturbed word-by-word. In this example, we employ the setting in which five word embedding models are used, with one list per model. An input word is \textit{diffracted} through these lists, producing a list of candidate perturbations, from which a final selection is made randomly.}
    \label{fig:overview}
\end{figure*}

The issue of data privacy has grown in relevance and attention in recent years with respect to the field of Natural Language Processing, particularly with the rising prominence of language models which require significant amounts of data to reach state-of-the-art performance \cite{9152761, carlini2021extracting}. Tasked with protecting individual privacy while also allowing for the continued proliferation of highly useful models, a number of solutions have appeared in recent literature to form the basis of \textit{privacy-preserving} Natural Language Processing.

One promising and increasingly researched solution comes with the notion of Differential Privacy (DP) \cite{Dwork2006DifferentialP}. In essence, Differential Privacy provides a mathematically grounded concept of individual privacy protection whose guarantees can be scaled according to the crucial $\varepsilon$ parameter, known as the \textit{privacy parameter}. However, the integration of DP into Natural Language Processing does not come without challenges \cite{feyisetan2021research, klymenko-etal-2022-differential, mattern-etal-2022-limits}, among them the transfer from structured data to textual data and reasoning about the \say{individual}. 

In response, one avenue of research looks to the \textit{word-level}, where text is obfuscated via word-by-word replacements, resulting in \textit{perturbed} data \cite{fernandes2019generalised, feyisetan_balle_2020}. These word-level methods often rely on \textit{Metric Local Differential Privacy} (MLDP), a generalized notion of DP which allows for the extension of DP into metric spaces, such as with words represented in embedding spaces \cite{chatzi_mdp_2013}. More recent works have made advancements in the selection strategy of the perturbed word \cite{carvalho2023tem}, distance metric \cite{feyisetan2019leveraging, xu2020differentially}, or calibration of mechanism according to the density of the embedding space \cite{xu2021density, xu2021utilitarian}.

As noted by \citet{mattern-etal-2022-limits}, a major shortcoming of word-level MLDP methods originates from the relatively large amounts of noise that must be added to satisfy DP, thus ultimately leading to perturbed textual data with poor utility. Another limitation, noted by \citet{klymenko-etal-2022-differential}, comes with the \say{structural limitations} imposed by MLDP, particularly when mapping from original text to perturbed text. Such perturbations require nearest neighbor searches, which can become computationally very expensive when working in high-dimensional spaces with large vocabularies.

In this work, we aim to address these two key issues of utility preservation and efficiency with word-level MLDP. To do so, we introduce 1-Dimensional Differentially Private Text Obfuscation (\textsc{1-Diffractor}), a novel method that is highly efficient and boasts competitive levels of utility preservation. In contrast to previous methods, \textsc{1-Diffractor} operates on single-dimensional word embedding lists and uses the geometric distribution, from which perturbation candidates are selected through what we call a \textit{diffraction} process. An illustration of this process is found in Figure \ref{fig:overview} and will be described in Section \ref{sec:method}.

To evaluate \textsc{1-Diffractor}, we set up three categories of experiments: (1) Utility Experiments, in which two versions of \textsc{1-Diffractor} are evaluated on the GLUE benchmark, (2) Privacy Experiments, which include a comparative analysis of our mechanism's privacy-preserving capabilities, as well as empirical privacy tests on two adversarial tasks, and (3) Efficiency Experiments, in which the performance and scalability our \textsc{1-Diffractor} is explored, particularly in comparison to previous mechanisms.

The results of our experiments demonstrate that perturbing datasets with \textsc{1-Diffractor} preserves utility across a variety of NLP tasks. In addition, \textsc{1-Diffractor} is effective in reducing adversarial advantage in two chosen tasks. Finally, \textsc{1-Diffractor} is significantly more efficient than previous methods, processing text at greater than 15x the speed and with less memory than previously.

The contributions of our work are as follows:

\begin{enumerate}
   \setlength\itemsep{0em}
    \item We present a novel word-level MLDP mechanism, built upon word embeddings in a one-dimensional space, or \textit{lists}
    \item We demonstrate the effectiveness of our list method with an existing noise-addition mechanism, as well as a new mechanism previously unused in the NLP domain
    \item We also emphasize \textit{efficient} word-level MLDP, highlighting the speed and memory consumption of word perturbations
\end{enumerate}

\section{Foundations}
\label{sec:foundations}
\subsection{Differential Privacy}
Intuitively, Differential Privacy (DP) \cite{Dwork2006DifferentialP} ensures that the result of a computation over a collection is nearly the same irrespective of inclusion or exclusion of a single data point. Hence, if we have two databases $\mathcal{D}$ and $\mathcal{D^\prime}$ differing in only one data point, when a differentially private mechanism is applied over these two databases $\mathcal{D}$ and $\mathcal{D^\prime}$, the result of the mechanism will be very similar. 
Such databases that differ only by a single element are called \textit{neighboring} or \textit{adjacent} databases. 
More formally, a mechanism $\mathcal{M}: \mathcal{X}^m \to \mathcal{O}$ operating over any two adjacent databases $\mathcal{D}$, $\mathcal{D^\prime} \in \mathcal{X}^m$ is $(\varepsilon,\delta)$-differentially private, iff $\forall O \subseteq \mathcal{O}$, the following condition holds:
\begin{displaymath}
    \mathbb{P}[\mathcal{M}(\mathcal{D}) \in O] 
     \leq e^\varepsilon \cdot {\mathbb{P}[\mathcal{M}(\mathcal{D^\prime}) \in O]} + \delta
\end{displaymath}
where $\varepsilon > 0$ and $\delta \in [0, 1]$. 

In other words, a mechanism is $(\varepsilon,\delta)$-differentially private if its output distributions on adjacent databases are "close enough" to each other. 
In the case of traditional databases, the notion of adjacent databases is simple to understand and without loss of generality could be given as $\mathcal{D^\prime} = \mathcal{D} \cup \{d\}$, where $d$ is a single record in a database. The case of unstructured domains such as text, however, brings additional considerations.
Based on how the notion of adjacent databases is defined, so is the element which DP aims to protect. Since this original notion defines databases as those differing in a single record, a differentially private mechanism $\mathcal{M}$ will guarantee that the influence of a single record on the output of the mechanism is bounded.

\subsection{Local Differential Privacy and NLP}
The notion of DP as introduced above is known as \textit{Global Differential Privacy}. Another notion is called \textit{Local Differential Privacy} (LDP), where noise is added directly to the data before being aggregated at a central location \cite{kasiviswanathan2011can}. In LDP, the notion of adjacent databases is defined over data points from a single individual: every collected data point from a single individual is adjacent to every other data point from another individual.
\citet{feyisetan_balle_2020} leverages the \say{\emph{one user, one word}} model, where the curator collects a word from each user and uses these to perform some downstream tasks. Instead of making the original words available to the analyst and leaking information about the users, users can run a privacy-preserving mechanism over their words before releasing them. 
In the base version of this model, the practicality is quite limited, as collecting a single word from each user would rarely allow for meaningful analysis. As noted by \citet{feyisetan_balle_2020}, however, this model can be extended to larger textual units (see Section \ref{sec:extend}).

Another aspect of this model is the notion of adjacency being defined on words -- each word is a database and is adjacent to every other word. However, this is a very strict notion of privacy with severe implications for utility \cite{feyisetan_balle_2020}. Hence, a relaxation of (LDP) called Metric Local DP (MLDP) or $d_\chi$-privacy \cite{chatzi_mdp_2013} is used instead for a better privacy-utility trade-off \cite{feyisetan_balle_2020}. 
 
\subsection{Metric (Local) Differential Privacy}
Let $\mathcal{X}$ and $\mathcal{Z}$ be finite sets and let $d: \mathcal{X} \times \mathcal{X} \to \mathbb{R+}$ be the distance metric defined on the set $\mathcal{X}$ that satisfies the axioms of a metric.

\begin{definition}
($d_\mathcal{X}$-privacy). Let $\varepsilon > 0$. A randomized mechanism $\mathcal{M}: \mathcal{X} \to \mathcal{Z}$ satisfies $\varepsilon d_\mathcal{X}$-privacy iff $\forall x, x^\prime \in \mathcal{X}$ and $\forall z \in \mathcal{Z}$

	\begin{equation}
		\frac{ \mathbb{P}[ \mathcal{M}(x) = z ] }
			 { \mathbb{P}[ \mathcal{M}(x^\prime) = z ] } 
		\leq e^{\varepsilon d(x, x^\prime)}
		\label{eq: edp}
	\end{equation}
\end{definition}

With MLDP, the notion of adjacent databases remains any two words, but while LDP bounds the output distributions over the adjacent sets by $e^\varepsilon$, MLDP bounds it by $e^{\varepsilon\cdot d_\chi(w, w^\prime)}$. Hence, when an MLDP mechanism is applied, the words that are close (measured by some distance metric) would have more \say{similar} output distributions compared to when the words are far apart. It should be noted that MLDP is a generalized notion of LDP, and (pure) LDP can be derived from MLDP by keeping the distance between any two words as a constant value, i.e., $\forall w, w^\prime \in \mathcal{X}, d_{\mathcal{X}}(w, w^\prime) = 1$.

Following \citet{feyisetan2} and subsequent works, these MLDP mechanisms are run on word \textit{embeddings}, which lend themselves well to the MLDP scheme due to their underlying metrics spaces and distance measures, while still preserving the goal of the \textit{one user, one word} notion.

\section{\textsc{1-Diffractor}}
\label{sec:method}
As previously stated, our model operates as defined by \citet{feyisetan_balle_2020}: \emph{one user, one word}. In this setting, a utility-preserving private mechanism produces a \say{perturbed} version of the input word that preserves its original intent but prevents the leakage of the user information that can be extracted from the choice of their word. If the word sent by a user is $w$, the mechanism $\mathcal{M}$ outputs its \say{privatized} word $\hat{w} = \mathcal{M}(w).$ Our Mechanism $\mathcal{M}$ operates on word embeddings where the words are arranged in a one-dimensional list, with adjacent words being close in the original space.

\paragraph{Intuition behind converting a word embedding model from $\mathbb{R}^d$ to $\mathbb{Z}$}
Previous word-level MLDP mechanisms operate on high-dimensional word embeddings, adding noise to every dimension of the vector. This not only adds a high value of noise (measured by its norm), but the noisy vector rarely corresponds to a word in the embedding model. Hence, it must be remapped to a nearby word, increasing the time complexity and potentially impacting the utility of the overall mechanism. There exist multiple approaches to remapping the vector to a word \cite{feyisetan_balle_2020, xu2021density, xu2021utilitarian}. Several approaches that do not add noise directly to word vectors have been proposed that use variants of the Exponential Mechanism for choosing a privatized word for the input word \cite{weggenmann2018syntf, carvalho2023tem}. We formulate our mechanism in a different way that is fast to compute, by reducing the dimension of embeddings to one dimension. As such, noise must only be added on one dimension; concretely, we add discretized noise sampled from Geometric distribution to words in one-dimensional space.

Converting high-dimensional embeddings to a 1-D list has several advantages: 1-D embeddings can be considered an index and this simplifies word privatization to returning a \textit{noisy index}. Moreover, these 1-D lists can be combined together into a collection of lists from different embedding models, thereby increasing the diversity of output words while also providing an extra layer of obfuscation, described further in Section \ref{sec:multiple}. 

To create such a one-dimensional list from a word embedding model such as word2vec \cite{mikolov2013efficient} or GloVe \cite{pennington-etal-2014-glove}, we initialize the list $L$ with a random word and iteratively add the nearest word in the embedding space to the previously added word. Concretely, a random word is first selected as the seed word and it is added to the list $L$. Then the nearest word in the embedding model, according to the Euclidean distance, is made the seed word and the process repeats for the remaining embedding space, until no words remain. 

Thus, this process mimics a \say{greedy search} through a given embedding space, from a randomly selected starting point. Algorithm \ref{alg: word_list} outlines this process. Multiple lists from a single embedding model can be created by initializing the starting point (seed) at different points. In addition, various pre-trained word embedding models can be utilized in tandem to create several lists.

\begin{algorithm}[htbp]
\caption{\newline Creation of a word list $L$ from a word embedding model}
\label{alg: word_list}
    \small
    \begin{algorithmic}
        \Require Word Embedding model $E$
        \State $L \gets \texttt{list}()$
        \State $\texttt{words} \gets \texttt{vocabulary}(E)$ 
        \State $\texttt{seed} \gets \texttt{random(words)}$
                    \State $L.\texttt{append(seed)}$
        \While {$\texttt{words.length()} > 1$}

            \State $N \gets \texttt{NearestWord(seed)}$
            \State $L.\texttt{append}(N)$
            \State $\texttt{words.remove(seed)}$
            \State $\texttt{seed} \gets N$
        \EndWhile
        \State \Return $L$
    \end{algorithmic}
\end{algorithm}

\subsection{Word-level $d_\mathcal{X}$-privacy mechanism}
\label{sec:geo}

\subsubsection{Using a single word embedding list} \label{ssect: mech_single_list}
We describe how our proposed $d_\chi$-privacy mechanism works with a single word embedding list $L$. We define an embedding function $\Phi: \mathcal{V} \to [0, |\mathcal{V}|] \cap \mathbb{N} $ over the list $L$ that takes a word from our vocabulary set $\mathcal{V}$ as input and returns its position (index) in the list as the output. Using these indices, we define a distance function $d_\mathcal{V}: \mathcal{V} \times \mathcal{V} \to \mathbb{Z}+$ that gives us the distance between two words $w$ and $w^\prime$ in our list as follows: 
\begin{equation}
    d_\mathcal{V}(w, w^\prime) = | \Phi(w) - \Phi(w') |
    \label{eq: distance}
\end{equation}

Note that the $d_\mathcal{V}$ distance function indeed follows all three axioms of a distance metric.
Now we define our mechanism that takes a word $w$ as input and outputs a \say{privatized} word $w^\prime$ using $d_\mathcal{V}$-privacy mechanism.

The mechanism $\mathcal{M}: \mathcal{V} \to \mathbb{Z}$ operates over a word and adds noise sampled from the geometric distribution. This particular distribution can be utilized due to our representation of words in one dimension (discrete indices), and thus we return a noisy index. Mathematically, $\mathcal{M}$ is defined as follows:
\begin{equation}
    \small
	\mathcal{M}(w, \Phi(\cdot), \varepsilon) = \Phi(w) + x,\\ x \sim \mathcal{G}\left(0,\frac{1}{\varepsilon}\right)
	\label{eq: proposed_mech}
\end{equation} 
where $\mathcal{G}$ is the Geometric distribution, given by the following probability density function,
\begin{equation}
    \small
        \forall x \in \mathbb{Z},\quad	\underset{X \gets \mathcal{G}(\mu, b) }{\mathbb{P}}[X = x] = \frac{ e^{1/b} - 1 }{ e^{1/b} + 1 } \cdot e^{ -\frac{|x - \mu|}{b} }
	\label{eq: gd}
\end{equation}

Using the property of linear transformation of random variables, one can see that our mechanism $\mathcal{M}$ is a randomized algorithm and its outputs are random variables drawn from the Geometric distribution $\mathcal{G}(\Phi(w), 1/\varepsilon)$. The proof that $\mathcal{M}$ satisfies $\varepsilon d_\mathcal{V}$-privacy can be found in Theorem \ref{the:dp_proof}.

\begin{theorem}
\label{the:dp_proof}
The proposed mechanism $\mathcal{M}$ defined in Equation \ref{eq: proposed_mech} satisfies $\varepsilon d_\mathcal{V}$-privacy.

	\begin{proof}
	Let $w$ and $w^\prime$ be any two words belonging to set $\mathcal{V}$, then the ratio of
	the probability distribution of application of $\mathcal{M}$ on $w$ and $w^\prime$
	can be given as
    \small
	\begin{align*}
		\frac{ \mathbb{P}[ \mathcal{M}(w) = x ] }
			 { \mathbb{P}[ \mathcal{M}(w^\prime) = x ] }  
		&=   \frac{ e^{-\varepsilon \cdot |x - \Phi(w)|} }
				  { e^{-\varepsilon \cdot |x - \Phi(w^\prime)|} }
		\\ & \quad\text{(From Equation \ref{eq: gd})}
		\\ 
		&= e ^ { \varepsilon \cdot (|x - \Phi(w^\prime)| - |x - \Phi(w)|) }
		\\ 
		&\leq e ^ { \varepsilon \cdot (|\Phi(w) - \Phi(w^\prime)|) }
		\\ & \text{($|a| - |b| \leq |a - b|$)}
		\\
		&= e ^ {\varepsilon \cdot d_V(w, w^\prime)}
		\\ & \quad\text{(From Equation \ref{eq: distance})}
	\end{align*}
	\end{proof}
\end{theorem}

Now, we define a truncation function $t: \mathbb{Z} \to [0, |\mathcal{V}|] \cap \mathbb{N}$ that takes the input from our mechanism defined above and truncates its output to the range $\{0 \dots |\mathcal{V}|\}$ in the following way:
\begin{equation}
    \small
	t(x) = 
	\begin{cases}
		x	&	x \in [0, |\mathcal{V}|] \\
		0	&	x < 0	\\
		
		|\mathcal{V}|	&	x > |\mathcal{V}|
	\end{cases}
\end{equation}

The application of the function $t(\cdot)$ to the output of the mechanism $\mathcal{M}$ truncates its values in the range $[0, |\mathcal{V}|]$. Due to resilience to post-processing of $\varepsilon d_\mathcal{X}$-privacy, the composition ($t \circ \mathcal{M}$) also satisfies $\varepsilon d_\mathcal{V}$-privacy \cite{Koufogiannis_Han_Pappas_2017}. Alternatively, the composition ($t \circ \mathcal{M}$) can be thought of as a randomized mechanism $\mathcal{M^\prime}$ that adds to the index $\Phi(w)$ a random variable $x$ drawn from a \textit{Truncated Geometric} distribution instead of the \textit{Geometric} distribution in Equation \ref{eq: proposed_mech}.

In order to convert the privatized index back to the domain of words, we can apply a function $r: [0, |\mathcal{V}|] \cap \mathbb{N} \to \mathcal{V}$. Note that function $r$ is the inverse function of our embedding function $\Phi(\cdot)$. $r(\cdot)$ takes the index of the word and returns its corresponding word. Again from the post-processing property of metric-DP, the composition ($r \circ t \circ \mathcal{M}$) satisfies $\varepsilon d_\mathcal{V}$-privacy \cite{Koufogiannis_Han_Pappas_2017}. Hence, our function $(r \circ t \circ \mathcal{M}): \mathcal{V} \to \mathcal{V}$ takes a word and outputs a ``privatized" word. 

\subsubsection{Using  multiple word embedding lists}
\label{sec:multiple}
One can use multiple lists $\mathbb{L} = \{L_1, L_2, \dots, L_n\}$ as well. We define separate word embedding functions $\Phi^l: \mathcal{V}\to [0, |\mathcal{V}|] \cap \mathbb{N}$ corresponding to each list $L_l$ and by extension, separate distance functions $d^l_{\mathcal{V}}: \mathcal{V} \times \mathcal{V} \to \mathbb{Z+}$ for each list $L_l$. Then, on the input word, we apply the mechanism  $\mathcal{M}(w, \Phi^l(\cdot), \varepsilon) $ for every list in $\mathbb{L}$, which outputs perturbed words $\mathbb{W} = \{w^\prime_1, w^\prime_2,\dots, w^\prime_n\}$ for an input word, and we \textbf{randomly} select a single word out of $\mathbb{W}$, releasing that as the \say{privatized} word for the input word. Even though the mechanism is applied $n$ times, only a single output out of all $n$ results is released, implying it is not a sequential application of $d_\chi$-privacy mechanism; hence, we do not incur any additional privacy cost in terms of $\varepsilon$ as compared to utilizing a single list. However, since the distance between the two words $w$ and $w^\prime$ may not be the same across lists, the probability distributions resulting from the application of mechanism $\mathcal{M}$ on $w$ and $w^\prime$ would be bounded by $e^{\varepsilon \cdot d_{max}(w, w^\prime)}$, i.e.,
\begin{equation}
    \small
    \frac{ \mathbb{P}[ \mathcal{M}(w) = x ] }
         { \mathbb{P}[ \mathcal{M}(w^\prime) = x ] } 
    \leq e^{\varepsilon d_{max}(w, w^\prime)}
\end{equation} 
where $d_{max}(w, w^\prime) = \max\limits_{l \in \{1 \dots n\}} d^l_{\mathcal{V}}(w, w^\prime)$.

\subsubsection{Extending Word-level $d_\mathcal{X}$-privacy to sentences}
\label{sec:extend}
Our proposed mechanism \textsc{1-Diffractor} operates on the word level; however, the perturbation of large units of textual data, i.e., sentences can be extrapolated from this base level, as described in \cite{feyisetan_balle_2020}.

In particular, a sentence $s$ can be considered as a concatenation of $n$ words, i.e., $s = w_1\cdot w_2\cdots w_n$. We follow \cite{feyisetan_balle_2020} and apply our mechanism to each word independently to generate a privatized sentence $s^\prime = x_1\cdot x_2 \cdots x_n$. We define the distance function $D: \mathcal{V^*} \times \mathcal{V^*} \to \mathbb{Z+}$ between two sentences of same length $s = w_1\cdots w_n$ and $s^\prime = w^\prime_1 \cdots w^\prime_n$ as 
$\mathcal{D} = \displaystyle\sum_{i=1}^n d_{max}(w_i, w^\prime_i)$..

Since the application of the mechanism $\mathcal{M}$ is independent with respect to words, when $\mathcal{M}$ is applied on a sentence $s = w_1\cdots w_n$, its output distribution is given by:
\begin{equation*}
    \small
    \mathbb{P}[\mathcal{M}(s) = z] = \displaystyle\prod_{i=1}^n \mathbb{P}[\mathcal{M}(w_i) = x]
\end{equation*}

\begin{theorem}
    $\forall s, s^\prime \in \mathcal{V^*}, \forall z \in \mathcal{V^*}$ and $|s| = |s^\prime| = |z| = n $, we have the following inequality: \\
    \small
    $$\frac{\mathbb{P}[\mathcal{M}(s) = z]}
         {\mathbb{P}[\mathcal{M}(s ^ \prime) = z]}
         \leq \exp{(\varepsilon \cdot \mathcal{D}(s, s^\prime))}$$
    \begin{proof}
        \begin{align*}
        \frac{\mathbb{P}[\mathcal{M}(s) = z]}
             {\mathbb{P}[\mathcal{M}(s^\prime) = z]}
        &=  \displaystyle\prod_{i=1}^n \frac{\mathbb{P}[\mathcal{M}(w_i) = x]}{\mathbb{P}[\mathcal{M}(w^\prime_i) = x]}
        \\
        &\leq \displaystyle\prod_{i=1}^n \exp{ \left( \varepsilon \cdot d_{max}(w_i, w_i^\prime) \right) }
        \\
        &=\exp{ \left( \varepsilon \cdot  \displaystyle\sum_{i=1}^nd_{max}(w_i, w_i^\prime) \right) }
        \\
        &=\exp{ \left( \varepsilon \cdot \mathcal{D}(s, s^\prime) \right)}
        \end{align*}
    \end{proof}
         
\end{theorem}
The limitation of extending word-level metric-DP to a sentence is that the neighboring \say{dataset} should be sentences of the same length, and hence the output sentence actually leaks the number of words in a sentence, as pointed out by \citet{mattern-etal-2022-limits}.

\section{Utility Experiments}
\label{sec:utility}
To test the utility-preservation of \textsc{1-Diffractor}, we have designed a three-part experiment, consisting of (1) experiments with different settings of our method on the \textsc{GLUE} benchmark, (2) comparative tests on selected \textsc{GLUE} tasks, comparing our method against previous MLDP mechanisms, and (3) a semantic similarity test to evaluate how well \textsc{1-Diffractor} preserves meaning.

\subsection{Design}
We decide to evaluate utility on the GLUE benchmark \cite{wang2019glue}, which presents a series of nine NLP tasks broken down into three categories. These include binary classification tasks (\textsc{CoLA}, \textsc{SST2}), textual similarity tasks (\textsc{QQP}, \textsc{MRPC}, \textsc{STSB}), and textual entailment tasks (\textsc{MNLI}, \textsc{QNLI}, \textsc{WNLI}, \textsc{RTE}). Previous works on text-to-text privatization perform evaluations on single tasks for the benchmark \cite{tang, yue}, but to the best of the authors' knowledge, no works have done so for the entire benchmark. Of particular note is the difficulty introduced when a benchmark dataset contains two sentences per data point, which presents an interesting test case for the utility-preserving capabilities of an MLDP mechanism. 

\subsubsection{Dataset Preparation}
For each dataset in the \textsc{GLUE} benchmark, we perturb all relevant columns with \textsc{1-Diffractor} in both the train and validation splits, i.e., either a single sentence or a sentence pair. For the larger datasets in the benchmark (\textsc{MNLI}, \textsc{QNLI}, \textsc{QQP}, \textsc{SST2}), we take 10\% of the training dataset. This is justified due to the large size of these datasets, as well as the fact that all experiments report \textit{relative} results to the baseline. In all cases, the full validation set is used. For \textsc{MNLI}, only the \textit{matched} split is used.

\subsubsection{Baseline Model and Scoring}
For all utility tests, we fine-tune BERT (\textsc{bert-base-uncased}) \cite{devlin-etal-2019-bert} on the train split, and report the evaluation performance on the validation split. For both the original dataset and all subsequent evaluations (perturbed datasets), the fine-tuning process is run on a V100 GPU (Google Colab) and is repeated \textbf{three} times, for one epoch each. This is to account for variations in the training process. Final scores are calculated by averaging the accuracy scores for the three runs, while also calculating the standard deviation. All tasks report accuracy scores, except for \textsc{STSB}, where the Pearson-Spearman Correlation is reported.

\subsubsection{Experiment Parameters}

\paragraph{Noise mechanism} We choose: (1) the \textit{Truncated Geometric} mechanism, as introduced in Section \ref{sec:geo}, denoted as $\textsc{1-D}_G$, and (2) the \textit{Truncated Exponential} mechanism (TEM), introduced by \citet{carvalho2023tem}, denoted as $\textsc{1-D}_T$. In the case of $\textsc{1-D}_T$, we adapt the usage of TEM for one-dimensional space, so as to fit \textsc{1-Diffractor}. In particular, $\textsc{1-D}_T$ operates on a subset of the vocabulary, governed by a truncation threshold $\gamma$, whose value depends on the chosen $\varepsilon$.

\paragraph{Choice of epsilon ($\varepsilon$)} We choose the values $\varepsilon \in \{0.1, 0.5, 1, 3, 5, 10\}$, which upon initial observation of our method, represents the \say{effective range} of noise addition, thus allowing for a test of strict privacy guarantees (e.g., 0.1) as well as weaker guarantees (e.g., 10). As will be noted in Section \ref{sec:epsilon}, this range is extended for comparative tests.

\paragraph{\textsc{1-Diffractor} settings} We utilize five embedding models\footnote{300-dim versions, except for E2 (200-dim)} and three list configurations:

\begin{itemize}
    \setlength\itemsep{0em}
    \item E1: \textit{conceptnet-numberbatch} \cite{speer2017conceptnet}    
    \item E2: \textit{glove-twitter} \cite{pennington-etal-2014-glove}
    \item E3: \textit{glove-wiki-gigaword} \cite{pennington-etal-2014-glove}
    \item E4: \textit{glove-commoncrawl} \cite{pennington-etal-2014-glove}
    \item E5: \textit{word2vec-google-news} \cite{mikolov2013efficient}

    \item \textbf{L0}: 1 list for each of E1-5
    \item \textbf{L1}: 2 lists for each of E1-5
    \item \textbf{L2}: 1 list, only E1
\end{itemize}

\subsection{Results}
\begin{figure}[htbp]
    \centering
    \includegraphics[scale=0.45]{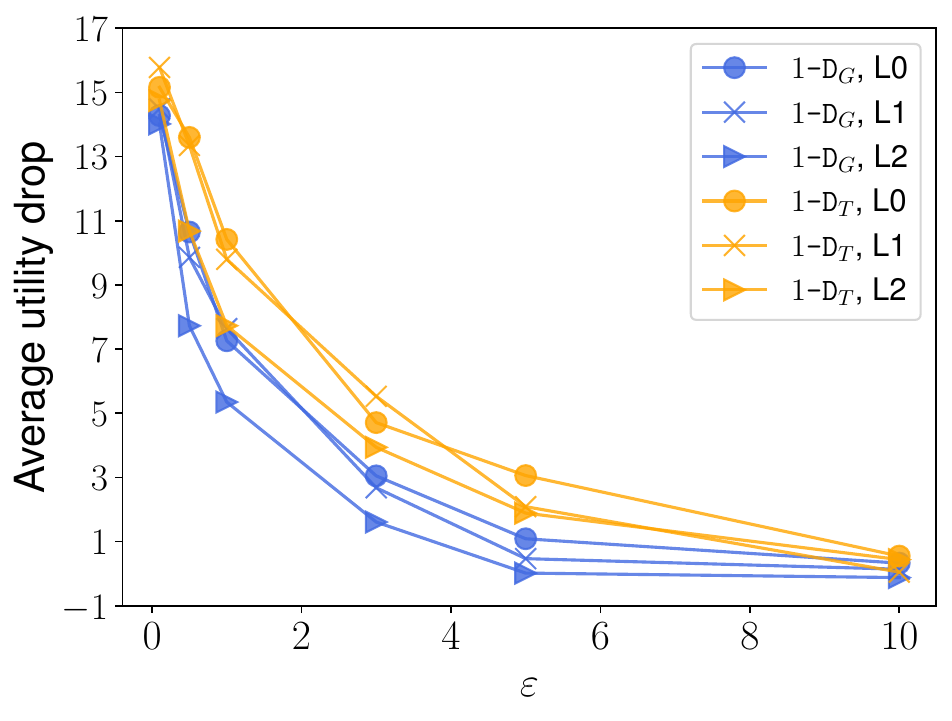} 
    \vspace{-1em}
    \caption{Average utility drop (loss) across all \textsc{GLUE} tasks of $\textsc{1-D}_G$ and $\textsc{1-D}_T$ with different list configurations and $\varepsilon$ values. Lower scores imply higher preserved utility.}
    \label{fig:utility_delta}
\end{figure}

\begin{figure}[htbp]
    \centering
    \includegraphics[scale=0.42]{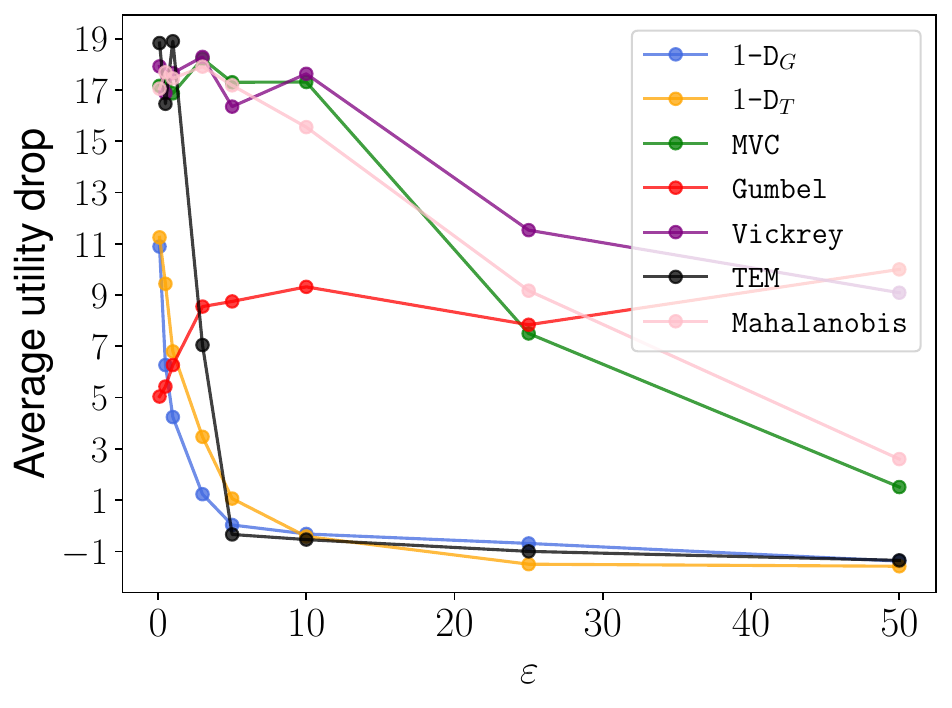}
    \vspace{-1em}
    \caption{Average utility drop across the \textsc{SST2}, \textsc{MRPC}, and \textsc{RTE} tasks compared to the five selected MLDP mechanisms.}
    \label{fig:utility_delta2}
\end{figure}

Table \ref{tab:utility} presents the full results of the \textsc{1-Diffractor} utility tests, for all of the above-mentioned experiment parameters. Figure \ref{fig:utility_delta} summarizes this information by illustrating the average utility drop (in percentage points) across all \textsc{GLUE} tasks, for each (\textit{mechanism}, $L$, $\varepsilon$) tuple. Similarly, Figure \ref{fig:utility_delta2} summarizes the utility results for the comparative test against the selected previous MLDP mechanisms. 

The full scores of \textsc{1-Diffractor} against the five selected MLDP mechanisms on three selected \textsc{GLUE} tasks can be found in Table \ref{tab:utility2}.

\begin{table*}[ht!]
    \begin{subtable}[h]{\linewidth}
    \centering
    \resizebox{\linewidth}{!}{%
        \begin{tabular}{ll|lll|lll||lll|lll|lll|}
         &  & \multicolumn{3}{c}{\textsc{CoLA}} & \multicolumn{3}{c||}{\textsc{SST2}} & \multicolumn{3}{c}{\textsc{QQP}} & \multicolumn{3}{c}{\textsc{MRPC}} & \multicolumn{3}{c|}{\textsc{STSB} (PCS)} \\ \cline{3-17} 
         & Baseline & \multicolumn{3}{c}{$80.57_{0.6}$} & \multicolumn{3}{c||}{$89.98_{0.5}$} & \multicolumn{3}{c}{$85.24_{0.2}$} & \multicolumn{3}{c}{$77.78_{0.8}$} & \multicolumn{3}{c|}{$88.17_{0.6}$} \\ \cline{3-17} 
         & $\varepsilon$ / L & \multicolumn{1}{c}{L0} & \multicolumn{1}{c}{L1} & \multicolumn{1}{c}{L2} & \multicolumn{1}{c}{L0} & \multicolumn{1}{c}{L1} & \multicolumn{1}{c||}{L2} & \multicolumn{1}{c}{L0} & \multicolumn{1}{c}{L1} & \multicolumn{1}{c}{L2} & \multicolumn{1}{c}{L0} & \multicolumn{1}{c}{L1} & \multicolumn{1}{c}{L2} & \multicolumn{1}{c}{L0} & \multicolumn{1}{c}{L1} & \multicolumn{1}{c|}{L2} \\ \hline
        \textsc{1-D}$_T$ & 0.1 &  $68.49_{0.7}$ & $68.01_{1.6}$ & $69.00_{0.2}$  & $69.30_{0.3}$ & $67.16_{0.5}$ & $69.50_{1.0}$ & $73.50_{0.6}$ & $73.40_{0.7}$ & $73.50_{0.7}$  & $68.55_{2.2}$ & $69.85_{1.2}$ & $70.02_{0.8}$  & $39.40_{5.1}$ & $41.42_{2.4}$ & $41.95_{1.8}$ \\
         & 0.5 &  $68.65_{0.4}$ & $68.94_{0.7}$ & $68.14_{1.1}$  & $74.73_{1.0}$ & $74.54_{1.4}$ & $76.45_{0.5}$ & $73.95_{1.4}$ & $73.79_{1.2}$ & $74.98_{0.7}$  & $69.12_{2.3}$ & $70.42_{0.2}$ & $72.39_{0.6}$  & $49.83_{2.4}$ & $49.67_{0.8}$ & $59.34_{0.8}$ \\
         & 1 & $68.36_{0.7}$ & $69.16_{0.0}$ & $70.63_{1.1}$  & $77.06_{0.2}$ & $76.99_{0.6}$ & $81.27_{0.7}$ & $77.24_{0.9}$ & $76.57_{0.3}$ & $78.27_{0.5}$  & $74.75_{1.4}$ & $72.55_{0.0}$ & $72.96_{0.8}$  & $59.38_{0.7}$ & $64.77_{1.1}$ & $69.38_{1.1}$  \\
         & 3 & $73.60_{0.5}$ & $72.61_{0.8}$ & $74.59_{0.2}$  & $85.93_{0.7}$ & $85.40_{0.3}$ & $88.07_{0.7}$ & $80.02_{0.6}$ & $80.40_{0.3}$ & $80.77_{0.5}$  & $72.88_{0.9}$ & $74.84_{0.7}$ & $73.12_{0.8}$  & $77.69_{0.4}$ & $79.00_{0.8}$ & $80.18_{0.8}$ \\
         & 5 &  $75.74_{1.0}$ & $77.21_{0.6}$ & $76.80_{0.5}$  & $88.23_{0.4}$ & $88.57_{0.5}$ & $89.79_{0.2}$ & $82.98_{0.2}$ & $82.40_{0.3}$ & $83.28_{0.3}$  & $75.82_{1.3}$ & $75.82_{1.0}$ & $75.41_{1.7}$  & $83.13_{0.6}$ & $83.65_{0.5}$ & $84.80_{0.7}$ \\
         & 10 &  $80.82_{0.4}$ & $81.43_{0.9}$ & $80.41_{0.2}$  & $89.60_{0.5}$ & $89.07_{0.2}$ & $89.22_{0.3}$ & $84.66_{0.5}$ & $84.31_{0.4}$ & $84.42_{0.2}$  & $78.84_{0.3}$ & $80.39_{0.5}$ & $77.29_{0.2}$  & $87.26_{0.5}$ & $87.52_{0.7}$ & $87.63_{0.6}$ \\ \hline \hline
         \textsc{1-D}$_G$ & 0.1 & $68.01_{1.2}$ & $67.95_{1.5}$ & $68.36_{0.5}$  & $69.30_{1.3}$ & $70.95_{0.5}$ & $71.75_{1.0}$ & $73.75_{0.5}$ & $74.42_{0.5}$ & $74.42_{0.5}$  & $70.34_{0.5}$ & $70.18_{1.2}$ & $68.71_{2.0}$  & $42.04_{2.4}$ & $44.33_{2.1}$ & $45.25_{1.8}$ \\
         & 0.5 &  $69.10_{1.1}$ & $69.22_{0.3}$ & $69.42_{0.4}$  & $77.10_{0.6}$ & $80.43_{0.6}$ & $81.46_{1.0}$ & $76.73_{0.5}$ & $74.92_{1.2}$ & $77.06_{0.6}$  & $70.59_{0.5}$ & $71.57_{1.0}$ & $71.08_{0.4}$  & $64.06_{1.7}$ & $63.40_{0.8}$ & $68.60_{0.7}$ \\
         & 1 &  $71.11_{0.3}$ & $70.69_{0.8}$ & $71.49_{0.3}$  & $82.95_{0.5}$ & $81.80_{0.1}$ & $85.40_{0.8}$ & $77.81_{0.7}$ & $78.65_{0.1}$ & $80.63_{0.1}$  & $75.90_{1.5}$ & $72.39_{0.6}$ & $74.02_{0.5}$  & $71.11_{0.7}$ & $73.24_{0.4}$ & $76.36_{1.2}$ \\
         & 3 &  $77.34_{0.6}$ & $77.05_{0.7}$ & $77.69_{0.4}$  & $88.19_{0.6}$ & $89.41_{0.7}$ & $89.79_{0.7}$ & $82.42_{0.4}$ & $82.46_{0.4}$ & $83.48_{0.3}$  & $75.74_{1.2}$ & $74.51_{1.0}$ & $77.21_{0.7}$  & $84.07_{0.5}$ & $83.11_{0.7}$ & $84.81_{0.4}$ \\
         & 5 &  $78.62_{0.9}$ & $79.45_{0.1}$ & $80.25_{0.3}$  & $89.41_{0.1}$ & $88.88_{1.1}$ & $89.37_{0.5}$ & $83.91_{0.4}$ & $84.03_{0.3}$ & $84.61_{0.4}$  & $78.10_{1.0}$ & $77.94_{0.7}$ & $78.92_{0.2}$  & $87.00_{0.5}$ & $87.06_{0.7}$ & $87.88_{0.6}$ \\
         & 10 & $81.08_{0.8}$ & $79.71_{0.4}$ & $80.98_{0.4}$  & $89.56_{0.5}$ & $88.99_{0.1}$ & $89.72_{0.4}$ & $83.79_{0.3}$ & $84.18_{0.1}$ & $84.88_{0.1}$  & $77.37_{0.8}$ & $77.86_{0.5}$ & $78.68_{0.2}$  & $87.07_{0.5}$ & $87.53_{0.8}$ & $88.14_{0.5}$ \\
        \end{tabular}
    }
    \caption{Utility Scores (Accuracy) for the Classification and Textual Similarity Tasks of GLUE. Note: for \textsc{STSB}, the (scaled) Pearson-Spearman Correlation (PCS) is given.}
    \label{tab:util1}
    \end{subtable}
    \hfill
    \begin{subtable}[h]{\linewidth}
    \centering
    \resizebox{0.95\linewidth}{!}{%
        \begin{tabular}{ll|lll|lll|lll|lll|}
         &  & \multicolumn{3}{c}{\textsc{MNLI}} & \multicolumn{3}{c}{\textsc{QNLI}} & \multicolumn{3}{c}{\textsc{WNLI}} & \multicolumn{3}{c|}{\textsc{RTE}} \\ \cline{3-14} 
         & Baseline & \multicolumn{3}{c}{$76.54_{0.5}$} & \multicolumn{3}{c}{$84.79_{0.5}$} & \multicolumn{3}{c}{$38.97_{2.4}$} & \multicolumn{3}{c|}{$59.21_{1.4}$} \\ \cline{3-14} 
         & $\varepsilon$ / L & \multicolumn{1}{c}{L0} & \multicolumn{1}{c}{L1} & \multicolumn{1}{c}{L2} & \multicolumn{1}{c}{L0} & \multicolumn{1}{c}{L1} & \multicolumn{1}{c}{L2} & \multicolumn{1}{c}{L0} & \multicolumn{1}{c}{L1} & \multicolumn{1}{c}{L2} & \multicolumn{1}{c}{L0} & \multicolumn{1}{c}{L1} & \multicolumn{1}{c|}{L2} \\ \hline
         \textsc{1-D}$_T$ & 0.1 &  $54.35_{1.6}$ & $54.18_{1.7}$ & $54.93_{1.1}$  & $67.98_{0.3}$ & $67.58_{0.8}$ & $69.08_{0.7}$  & $48.83_{3.3}$ & $41.31_{1.8}$ & $46.01_{1.8}$  & $54.39_{3.7}$ & $56.32_{0.5}$ & $54.51_{1.1}$ \\
         & 0.5 & $56.87_{1.3}$ & $57.36_{1.1}$ & $60.33_{1.0}$  & $69.18_{1.0}$ & $71.33_{0.5}$ & $72.99_{0.8}$  & $44.13_{5.8}$ & $42.72_{2.9}$ & $46.95_{2.4}$  & $52.35_{0.6}$ & $52.35_{0.8}$ & $53.55_{4.7}$  \\
         & 1 &  $61.27_{0.6}$ & $60.85_{1.2}$ & $63.57_{0.9}$  & $73.45_{0.6}$ & $72.99_{1.2}$ & $76.94_{0.3}$  & $42.72_{2.4}$ & $44.13_{3.3}$ & $42.72_{2.9}$  & $53.19_{1.8}$ & $54.99_{1.3}$ & $55.96_{1.5}$ \\
         & 3 & $68.37_{1.0}$ & $68.23_{0.9}$ & $71.19_{0.4}$  & $80.82_{0.6}$ & $78.43_{0.8}$ & $79.41_{0.8}$  & $43.19_{1.3}$ & $37.56_{3.7}$ & $40.38_{2.4}$  & $56.32_{0.8}$ & $54.99_{0.3}$ & $58.12_{0.6}$ \\
         & 5 &  $72.77_{0.4}$ & $72.78_{0.5}$ & $74.12_{0.2}$  & $80.37_{0.4}$ & $82.31_{0.3}$ & $81.16_{0.3}$  & $36.15_{2.4}$ & $39.44_{2.3}$ & $39.91_{1.3}$  & $58.48_{0.6}$ & $60.29_{2.8}$ & $58.97_{2.1}$  \\
         & 10 &  $76.43_{0.3}$ & $75.72_{0.6}$ & $76.71_{0.3}$  & $83.33_{0.0}$ & $84.06_{0.2}$ & $82.79_{0.6}$  & $36.15_{4.4}$ & $38.03_{4.1}$ & $38.03_{1.1}$  & $59.21_{0.8}$ & $60.29_{1.0}$ & $60.77_{1.6}$ \\ \hline \hline
         \textsc{1-D}$_G$ & 0.1 & $55.85_{1.3}$ & $54.75_{1.3}$ & $56.60_{1.3}$  & $66.90_{0.9}$ & $68.41_{0.7}$ & $69.67_{1.1}$  & $52.58_{2.9}$ & $47.42_{7.7}$ & $45.07_{1.1}$  & $53.91_{3.2}$ & $52.47_{3.0}$ & $55.23_{1.9}$\\
         & 0.5 &  $59.90_{1.2}$ & $61.34_{0.8}$ & $63.91_{1.0}$  & $72.71_{0.9}$ & $71.89_{1.0}$ & $74.18_{0.4}$  & $41.31_{4.6}$ & $40.38_{2.4}$ & $46.95_{6.3}$  & $53.91_{2.5}$ & $59.33_{2.4}$ & $58.97_{0.6}$  \\
         & 1 & $64.72_{1.1}$ & $65.80_{0.9}$ & $67.98_{0.8}$  & $74.62_{1.0}$ & $76.89_{1.0}$ & $80.30_{0.7}$  & $38.97_{0.7}$ & $38.50_{2.4}$ & $39.91_{4.8}$  & $58.72_{1.2}$ & $54.51_{1.5}$ & $57.04_{2.1}$  \\
         & 3 &  $73.47_{0.7}$ & $73.83_{0.5}$ & $75.05_{0.5}$  & $80.61_{0.7}$ & $81.76_{0.4}$ & $82.37_{0.7}$  & $35.21_{3.0}$ & $35.21_{1.1}$ & $38.03_{2.3}$  & $56.80_{0.9}$ & $59.81_{0.9}$ & $58.36_{1.2}$\\
         & 5 & $75.54_{0.5}$ & $76.10_{0.4}$ & $76.79_{0.6}$  & $82.00_{0.7}$ & $84.36_{0.3}$ & $83.04_{0.7}$  & $38.50_{1.3}$ & $40.38_{1.3}$ & $39.44_{2.0}$  & $58.36_{0.2}$ & $58.84_{0.5}$ & $60.77_{0.9}$  \\
         & 10 & $76.17_{0.3}$ & $76.49_{0.3}$ & $77.13_{0.4}$  & $85.17_{0.2}$ & $85.01_{0.1}$ & $84.97_{0.4}$  & $38.03_{4.0}$ & $38.50_{3.7}$ & $38.03_{4.0}$  & $60.05_{0.3}$ & $61.73_{1.3}$ & $59.81_{0.5}$ \\ 
        \end{tabular}%
    }
        \caption{Utility Scores (Accuracy) for the Textual Entailment Tasks of GLUE.}
        \label{tab:util2}
    \end{subtable}
    \caption{Utility Scores for \textsc{1-Diffractor} across the nine GLUE tasks, for six selected $\varepsilon$ values. $L$ values denote the three different list configuration settings that were used for the utility experiments. Scores presented are an average of three separate training runs, and the standard deviation is also presented for each score (as a subscript).}
    \label{tab:utility}
    \vspace{-1em}
\end{table*}

\begin{table}[htbp]
\resizebox{\linewidth}{!}{
\begin{tabular}{l|cccccccc}
\multicolumn{1}{r}{$\varepsilon$} & 0.1 & 0.5 & 1 & 3 & 5 & 10 & 25 & 50\\ \hline
$1-D_G$ & \textbf{0.37} & \textbf{0.57} & \textbf{0.66} & \textbf{0.80} & \textbf{0.82} & \textbf{0.82} &  0.99 & 0.99 \\
$1-D_T$ & 0.31 & 0.48 & 0.57 & 0.73 & 0.79 & \textbf{0.82} & \textbf{1.00} & \textbf{1.00} \\ \hline \hline
MVC & 0.08 & 0.08 & 0.08 & 0.09 & 0.10 & 0.15 & 0.64 & 0.77 \\
Gumbel & 0.33 & 0.33 & 0.33 & 0.33 & 0.33 & 0.33 & 0.59 & 0.61 \\
Vickrey & 0.08 & 0.08 & 0.08 & 0.09 & 0.10 & 0.14 & 0.42 & 0.49 \\
TEM & 0.10 & 0.10 & 0.11 & 0.38 & 0.45 & 0.45 & 0.83 & 0.83 \\
\small Mahalanobis & 0.09 & 0.08 & 0.09 & 0.09 & 0.10 & 0.14 & 0.56 & 0.75
\end{tabular}
}
\caption{Average SBERT cosine similarity scores for (original, perturbed) sentence pairs of the MRPC, RTE, and SST2 tasks. \textsc{1-D} variants are averaged for all list configurations (L0-L2). Best scores for each $\varepsilon$ value are bolded.}
\label{tab:sbert}
\end{table}

\begin{table*}[htbp]
    \small
    \centering
\begin{tabular}{l|lllllllll}
\toprule
 & \multicolumn{1}{r|}{Mechanism / $\varepsilon$} & \multicolumn{1}{c}{0.1} & \multicolumn{1}{c}{0.5} & \multicolumn{1}{c}{1} & \multicolumn{1}{c}{3} & \multicolumn{1}{c}{5} & \multicolumn{1}{c}{10} & \multicolumn{1}{c}{25} & \multicolumn{1}{c}{50}\\ 
\hline
\multirow{6}*{SST2} & \multicolumn{1}{l|}{1-D Best} & 71.75 $\pm$ 1.0 & \textbf{81.46 $\pm$ 1.0} & \textbf{85.40 $\pm$ 0.8} & \textbf{89.79 $\pm$ 0.7} & \textbf{89.79 $\pm$ 0.2} & \textbf{89.72 $\pm$ 0.4} & 89.72 $\pm$ 0.6 & 89.72 $\pm$ 0.6 \\
\cline{2-10}
& \multicolumn{1}{l|}{MVC} & 53.78 $\pm$ 2.0 & 55.01 $\pm$ 3.0 & 52.98 $\pm$ 1.5 & 54.32 $\pm$ 2.4 & 57.42 $\pm$ 1.2 & 59.52 $\pm$ 0.8 & 80.05 $\pm$ 0.8 & 86.23 $\pm$ 0.8\\
& \multicolumn{1}{l|}{Gumbel} & \textbf{78.44 $\pm$ 0.4} & 77.06 $\pm$ 0.1 & 75.27 $\pm$ 1.4 & 76.72 $\pm$ 1.4 & 76.61 $\pm$ 1.9 & 76.61 $\pm$ 1.2 & 78.36 $\pm$ 1.4 & 76.41 $\pm$ 0.2 \\
& \multicolumn{1}{l|}{Vickrey} & 53.25 $\pm$ 1.7 & 55.16 $\pm$ 0.7 & 53.90 $\pm$ 2.1 & 53.59 $\pm$ 2.0 & 56.96 $\pm$ 0.8 & 56.88 $\pm$ 0.8 & 71.67 $\pm$ 0.5 & 74.16 $\pm$ 1.0 \\
& \multicolumn{1}{l|}{TEM} & 53.13 $\pm$ 0.8 & 54.97 $\pm$ 2.9 & 51.34 $\pm$ 0.7 & 78.10 $\pm$ 0.4 & 89.26 $\pm$ 0.4 & 89.14 $\pm$ 1.5 & \textbf{90.14 $\pm$ 0.5} & \textbf{90.21 $\pm$ 1.1} \\
& \multicolumn{1}{l|}{Mahalanobis} & 54.74 $\pm$ 3.0 & 55.01 $\pm$ 2.4 & 54.01 $\pm$ 2.2 & 53.78 $\pm$ 1.5 & 53.82 $\pm$ 1.4 & 59.40 $\pm$ 0.7 & 74.89 $\pm$ 0.4 & 85.32 $\pm$ 0.4 \\ 
\hline \hline

\multirow{6}*{RTE} 
& \multicolumn{1}{l|}{1-D Best} & \textbf{56.32 $\pm$ 0.5} & \textbf{59.33 $\pm$ 2.4} & \textbf{58.72 $\pm$ 1.2} & \textbf{59.81 $\pm$ 0.9} & \textbf{60.77 $\pm$ 0.9} & \textbf{61.73 $\pm$ 1.3} & \textbf{65.00 $\pm$ 3.6} & \textbf{63.70 $\pm$ 2.7} \\ 
\cline{2-10}
& \multicolumn{1}{l|}{MVC} & 52.59 $\pm$ 1.0 & 52.47 $\pm$ 1.6 & 54.51 $\pm$ 1.0 & 50.42 $\pm$ 2.5 & 48.38 $\pm$ 1.1 & 48.86 $\pm$ 1.9 & 55.20 $\pm$ 0.0 & 59.81 $\pm$ 3.3 \\
& \multicolumn{1}{l|}{Gumbel} & \textbf{56.32 $\pm$ 2.3} & 54.39 $\pm$ 0.3 & 54.99 $\pm$ 2.2 & 53.67 $\pm$ 1.6 & 52.95 $\pm$ 1.2 & 52.71 $\pm$ 2.1 & 54.99 $\pm$ 0.7 & 51.26 $\pm$ 1.8 \\
& \multicolumn{1}{l|}{Vickrey} & 51.99 $\pm$ 0.6 & 54.15 $\pm$ 2.5 & 51.62 $\pm$ 3.1 & 51.26 $\pm$ 2.7 & 52.59 $\pm$ 0.5 & 49.22 $\pm$ 1.5 & 53.79 $\pm$ 0.8 & 55.60 $\pm$ 0.3\\
& \multicolumn{1}{l|}{TEM} & 49.94 $\pm$ 2.7 & 53.19 $\pm$ 2.2 & 51.87 $\pm$ 3.4 & 58.84 $\pm$ 1.9 & 59.81 $\pm$ 1.0 & 60.29 $\pm$ 1.4 & 61.50 $\pm$ 3.3 & 62.21 $\pm$ 3.8\\
& \multicolumn{1}{l|}{Mahalanobis} & 53.43 $\pm$ 1.5 & 51.50 $\pm$ 0.9 & 51.99 $\pm$ 1.2 & 52.23 $\pm$ 1.2 & 51.99 $\pm$ 1.6 & 53.43 $\pm$ 1.8 & 54.63 $\pm$ 2.0 & 59.57 $\pm$ 1.3\\
\hline \hline

\multirow{6}*{MRPC} 
& \multicolumn{1}{l|}{1-D Best} & 70.34 $\pm$ 0.5 & \textbf{72.39 $\pm$ 0.6} & \textbf{75.90 $\pm$ 1.5} & \textbf{77.21 $\pm$ 0.7} & \textbf{78.92 $\pm$ 0.2} & \textbf{80.39 $\pm$ 0.5} & \textbf{79.24 $\pm$ 0.7} & \textbf{79.41 $\pm$ 0.9} \\ 
\cline{2-10}
& \multicolumn{1}{l|}{MVC} & 69.12 $\pm$ 0.6 & 67.89 $\pm$ 1.0 & 68.87 $\pm$ 0.3 & 67.57 $\pm$ 1.3 & 69.28 $\pm$ 0.4 & 66.67 $\pm$ 1.7 & 69.20 $\pm$ 0.7 & 76.39 $\pm$ 0.9 \\
& \multicolumn{1}{l|}{Gumbel} & \textbf{70.59 $\pm$ 0.9} & 71.73 $\pm$ 0.5 & 69.44 $\pm$ 1.5 & 70.92 $\pm$ 0.6 & 71.16 $\pm$ 0.8 & 69.69 $\pm$ 1.4 & 70.10 $\pm$ 0.6 & 69.28 $\pm$ 0.2 \\
& \multicolumn{1}{l|}{Vickrey} & 67.97 $\pm$ 3.2 & 66.99 $\pm$ 1.3 & 68.46 $\pm$ 0.5 & 67.24 $\pm$ 0.8 & 68.38 $\pm$ 1.1 & 67.97 $\pm$ 0.6 & 66.91 $\pm$ 1.0 & 69.93 $\pm$ 0.7 \\
& \multicolumn{1}{l|}{TEM} & 67.40 $\pm$ 2.3 & 69.44 $\pm$ 0.8 & 67.08 $\pm$ 2.7 & 68.87 $\pm$ 3.7 & \textbf{78.92 $\pm$ 0.5} & 79.17 $\pm$ 0.3 & 78.35 $\pm$ 0.6 & 78.59 $\pm$ 0.3 \\
& \multicolumn{1}{l|}{Mahalanobis} & 67.65 $\pm$ 3.0 & 67.40 $\pm$ 3.0 & 68.63 $\pm$ 0.9 & 67.24 $\pm$ 2.0 & 69.61 $\pm$ 1.0 & 67.48 $\pm$ 0.7 & 69.93 $\pm$ 0.5 & 74.26 $\pm$ 1.2\\
\bottomrule
\end{tabular}

\caption{Utility Experiment Results with previous MLDP mechanisms, on three selected \textsc{GLUE} tasks. Scores represent the average of three runs, and standard deviations are presented. \textit{1-D Best} denotes the highest score achieved by a \textsc{1-Diffractor} configuration, i.e., (\textit{mechanism}, $L$) pair, from all scores presented in Table \ref{tab:utility}. \textbf{Bolded} values represent the best scores per $\varepsilon$ value.}
\label{tab:utility2}
\end{table*}

\paragraph{SBERT}
As a final part of our conducted utility experiments, we employ an SBERT model \cite{reimers-gurevych-2019-sentence}, namely \texttt{all-MiniLM-L6-v2}, to compare the semantic similarity between original and privatized sentences. This approach is similar to that proposed in BERTScore \cite{bert-score} for evaluating text generation. As noted by \citet{mattern-etal-2022-limits}, this metric is important for evaluating the ability of a text obfuscation mechanism to preserve the semantic coherence of the original sentence. The scores, grouped by $\varepsilon$ value, are provided in Table \ref{tab:sbert}.

\section{Privacy Experiments}
\label{sec:privacy}
To evaluate the privacy-preservation capabilities of \textsc{1-Diffractor}, we employ a two-fold approach. Firstly, the theoretical privacy guarantees of the model are tested against previous MLDP mechanisms. Next, empirical privacy tests are run on two adversarial tasks to evaluate the ability of \textsc{1-Diffractor} to obfuscate text.

\subsection{Plausible Deniability}
Following Feyistan et al. \cite{feyisetan_balle_2020}, we calculate plausible deniability statistics for our mechanism, which provide an idea of the variation introduced by a given text perturbation mechanism, for a given $\varepsilon$ value. In particular, there are two statistics: $N_w$, which measures the probability that a word is returned unperturbed (i.e., is perturbed to itself), and $S_w$, which measures the \textit{support} of perturbing a certain word, i.e, how many output words are expected given an input word. To estimate $N_w$ and $S_w$, we randomly sample 100 words from the vocabulary of models E1-5. Using list configuration L0, each of the 100 words is perturbed 100 times through \textsc{1-Diffractor}.

\begin{figure}[htbp]
    \centering
    \begin{subfigure}[b]{\linewidth}
        \centering
        \includegraphics[scale=0.4]{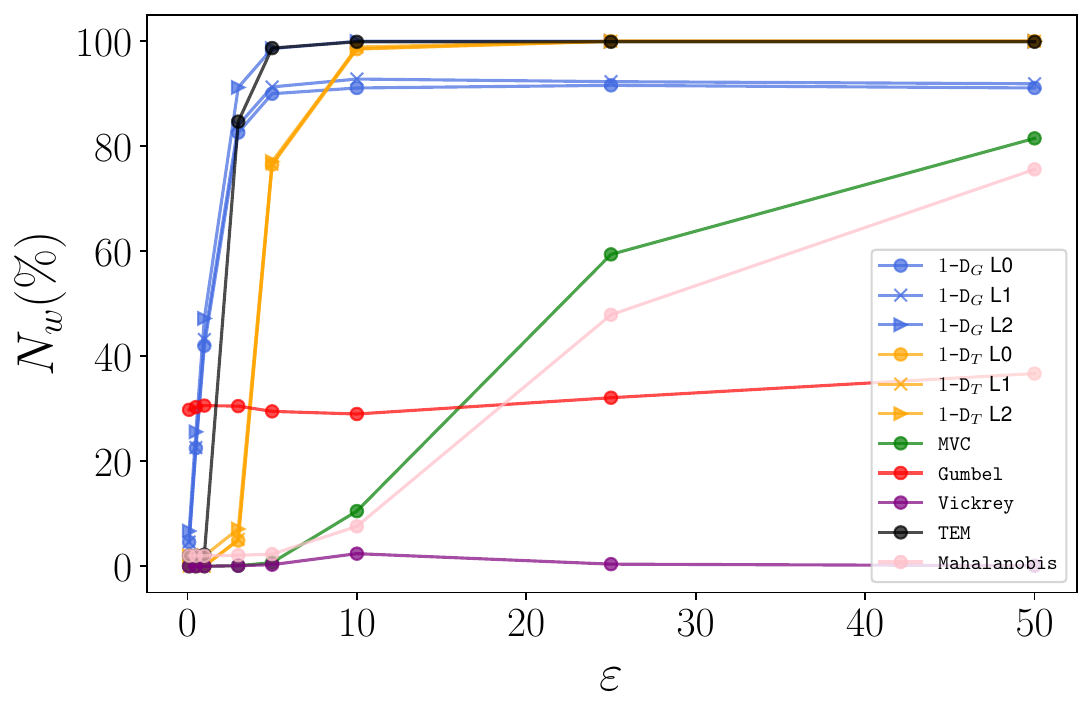}
        \label{fig:n_w}
    \end{subfigure}
  \begin{subfigure}[b]{\linewidth}
        \centering
        \includegraphics[scale=0.4]{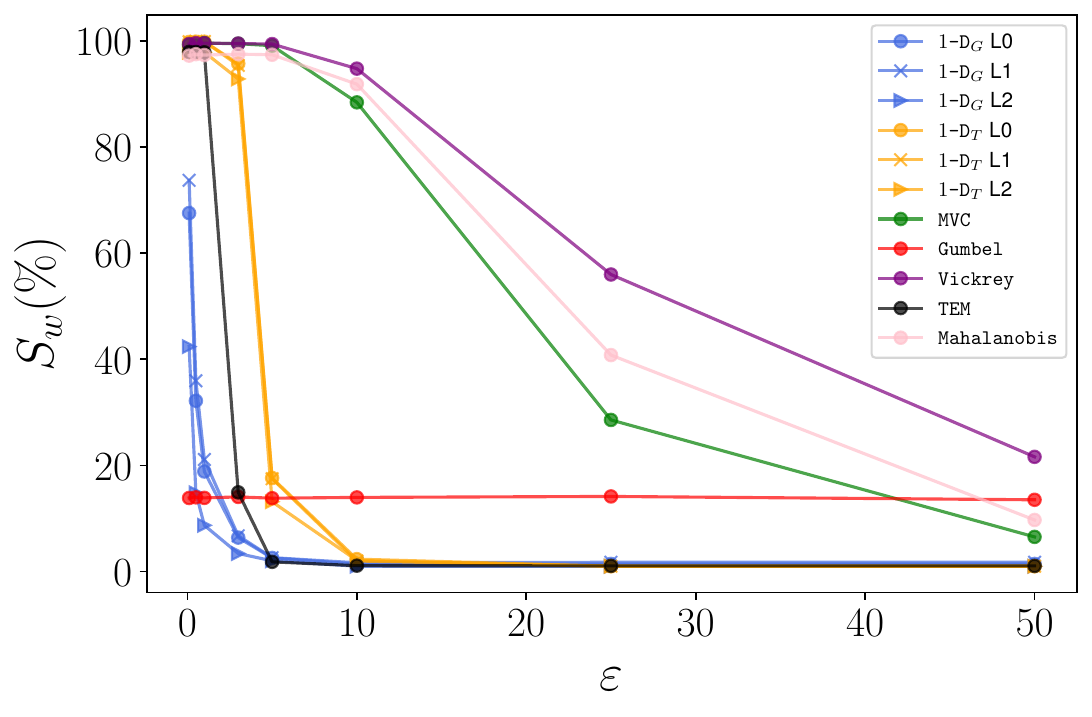}
        \label{fig:s_w}
    \end{subfigure}
  \caption{Empirical $N_w$ and $S_w$ statistics for \textsc{1-Diffractor} and five selected MLDP mechanisms.}
  \label{fig:nwsw}
\end{figure}

Figure \ref{fig:nwsw} displays the results of these tests. For comparison, we also test five recent MLDP mechanisms, namely: \texttt{MVC} \cite{feyisetan_balle_2020}, \texttt{Gumbel} \cite{xu2021density}, \texttt{Vickrey} \cite{xu2021utilitarian}, \texttt{TEM} \cite{carvalho2023tem}, and \texttt{Mahalanobis} \cite{xu2020differentially}, the set of mechanisms used for comparative testing in the remainder of this work.

\subsection{Empirical Privacy}
\begin{figure}[htbp]
    \centering
    \begin{subfigure}{0.8\linewidth} 
    \centering
    \includegraphics[width=\textwidth]{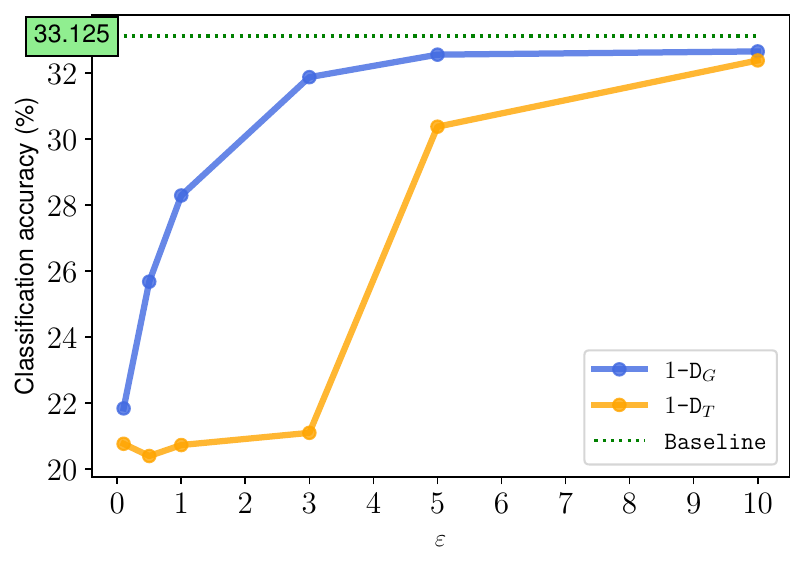}
    \vspace{-2em}
    \caption{\textbf{FI} results.}
    \label{fig:ep_fi}
  \end{subfigure}
  \hfill
  \begin{subfigure}{0.8\linewidth}
    \centering
    \includegraphics[width=\textwidth]{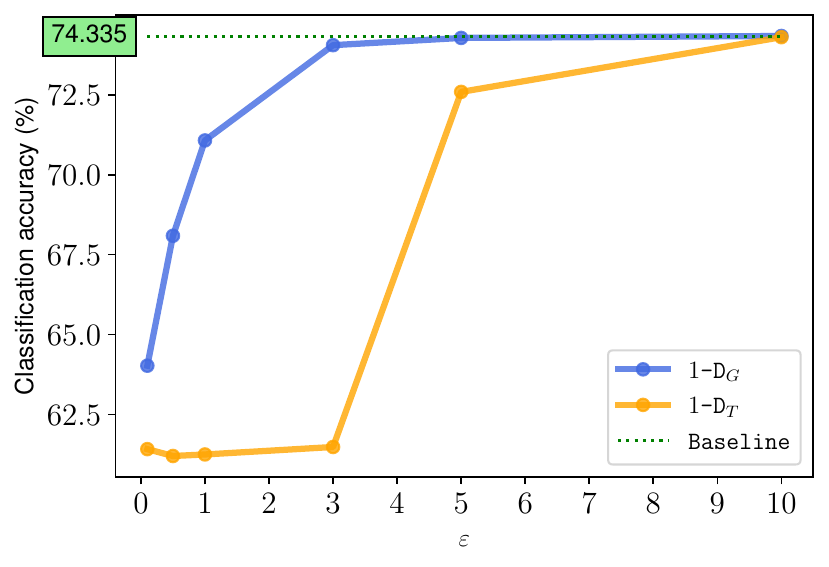}
    \vspace{-2em}
    \caption{\textbf{TG} results.}
    \label{fig:ep_tg}
  \end{subfigure}
  \caption{Emprical Privacy Results. \textit{FI} = Friends identification task, \textit{TG} = Trustpilot gender task.}
  \label{fig:ep}
\end{figure}

For empirical privacy tests, we choose two tasks: \textit{speaker identification} and \textit{gender identification}. The results are visualized in Figure \ref{fig:ep} and presented in full in Table \ref{tab:results_ep}.

In the speaker identification task (\textbf{FI}), we use the \textit{Friends Corpus} \cite{chen-choi-2016-character}, which contains the entirety of the script from the TV show \textit{Friends}. We take a subset of only the six main characters, and fine-tune a BERT model to identify a character based on their line. In the adversarial setting, this model mimics an attacker who wishes to identify authors based upon publicly accessible textual data. 

In the second task (\textbf{TG}), we use a dataset of US-based reviews on \textit{Trustpilot} \cite{10.1145/2736277.2741141}. Each review has been marked with the gender of the author. From this, we fine-tune BERT to predict the gender of an author based on the review text.

In both cases, the model acting as our adversary is trained with an 80\% split of the dataset, using a 10\% validation set. The 10\% test set is used to obtain the baseline scores for the adversarial classifier. Next, the test set is perturbed using \textsc{1-Diffractor} with the mentioned $\varepsilon$ values, using the \textbf{L0} configuration. Finally, the adversarial accuracy is evaluated for each perturbed dataset.

\begin{table}[htbp]
\centering
    \scriptsize
    \resizebox{\linewidth}{!}{
    \begin{tabular}[width=\textwidth]{l|r|cccccc}
            &   $\varepsilon$  &   0.1 &   0.5 &   1   &   3   &   5   &   10  \\
        \hline
        \multirow{3}{*}{\textbf{FI}} &   \textit{Baseline}    &   \multicolumn{6}{c}{33.13} \\
            &   $\textsc{1-D}_G$   &   21.84  &   25.68  &   28.29  &   31.88  &   32.56  &   32.66 \\
            &   $\textsc{1-D}_T$ &   20.77  &   20.40  &   20.73  &   21.10  &   30.38  &   32.39 \\
            \hline \hline
         \multirow{3}{*}{\textbf{TG}} &   \textit{Baseline}    &   \multicolumn{6}{c}{74.34} \\
            &   $\textsc{1-D}_G$   &   64.03  &   68.09  &   71.07  &   74.06 &   74.28  &   74.34 \\
            &   $\textsc{1-D}_T$ &   61.42  &   61.20  &   61.25  &   61.48  &   72.59  &   74.31  \\

       \bottomrule
    \end{tabular}
    }
    \caption{Complete empirical privacy results (accuracy). \textit{FI} = Friends identification task, \textit{TG} = Trustpilot gender task.}
    \label{tab:results_ep}
\end{table}

\subsection{A Note on Epsilon}
\label{sec:epsilon}
In the comparative analysis presented, it is important to note that the choice of $\varepsilon$ does not necessarily equate to the same privacy guarantees across mechanisms. This is due to different distance metrics being used, which scale the chosen $\varepsilon$ according to MDP. Therefore, the $\varepsilon$ values in our experiments are chosen for comparison, not necessarily to equate the \textit{effective} $\varepsilon$ values. Nevertheless, we mitigate this challenge by also testing all mechanisms including \textsc{1-Diffractor} on $\varepsilon \in \{25, 50\}$ in the comparative evaluations, thereby extending the investigated range. These results can be found in Tables \ref{tab:sbert} and \ref{tab:utility2}, as well as in Figures \ref{fig:utility_delta2} and \ref{fig:nwsw}.

\section{Efficiency Experiments}
\label{sec:speed}
The final experiments aim to measure the scalability of our proposed \textsc{1-Diffractor} mechanism, in comparison to previous MLDP mechanisms. This is performed by measuring the \textit{speed} and \textit{memory consumption} of each mechanism, quantified by the number of tokens that can be perturbed in a day and memory usage per word perturbation, respectively. Note that in these calculations, the list initialization of \textsc{1-Diffractor} is not included, as the time is negligible (ca. 20 seconds per list on a CPU) with respect to the 24-hour period, and 150 MiB for the initialization of \textbf{L0}.

We first estimate efficiency by capturing the amount of time it takes to perturb a random set of 1000 words from the list vocabulary. Next, we measure efficiency empirically by using each mechanism to perturb the complete \textsc{SST2} dataset. The number of perturbed tokens is divided by the elapsed and then extrapolated to 24 hours.

The second set of experiments focuses on the memory consumption of the word perturbations. To measure this, we measure the memory needed to perturb the same set of 1000 words as introduced above, using the Python \texttt{memory-profiler} package.

\begin{figure}[ht!]
    \centering
    \includegraphics[scale=0.5]{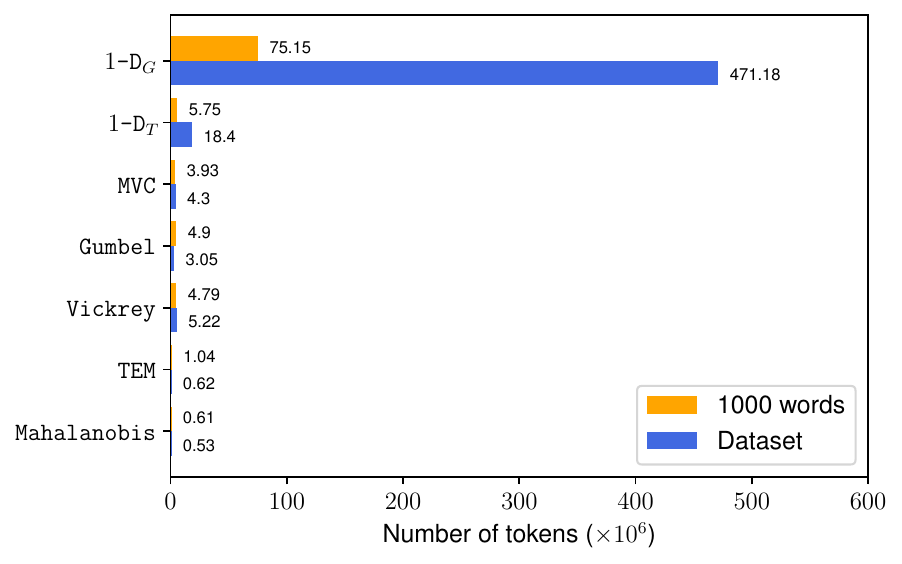} 
    \caption{Comparison of number of tokens that can be processed per day by calculating the speed over 1000 words and extrapolating to 24 hours (denoted by \textit{1000 words}). \textit{Dataset} is derived from the perturbation of the SST2 dataset.}
    \label{fig:speed}
\end{figure}

\begin{figure}[htbp]
    \centering
    \includegraphics[scale=0.29]{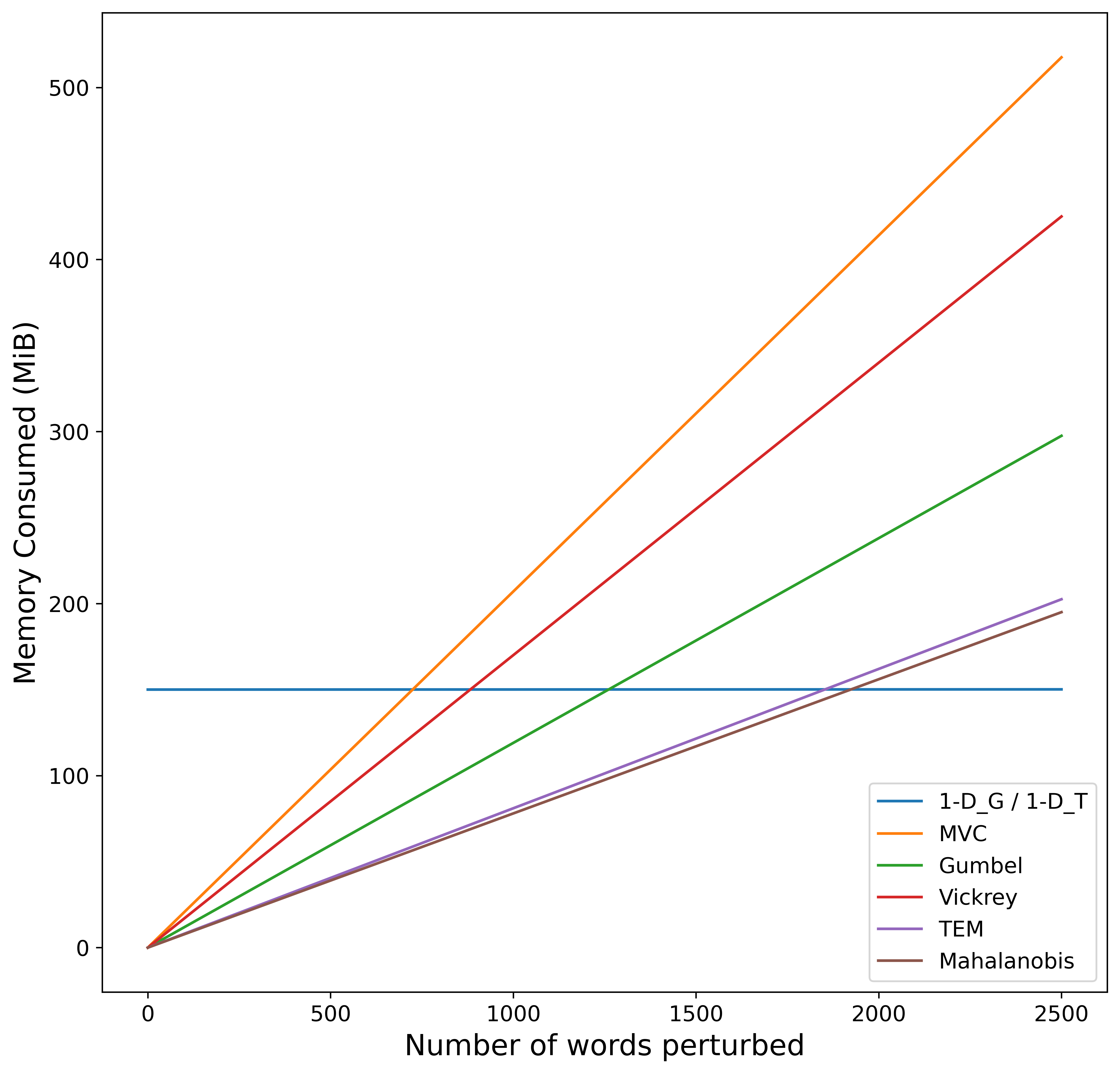} 
    \caption{Memory consumption of all compared mechanisms as a function of words perturbed, using the per-word rates of Table \ref{tab:memory}. For the \textsc{1-Diffractor} mechanisms, the memory required for initial list configuration is accounted for (see Section \ref{sec:speed}). In addition, the two \textsc{1-Diffractor} variants are plotted on the same line, due to their negligibly small difference in memory required. Note that these plots assume a linear growth in memory consumption over time.}
    \label{fig:memory}
\end{figure}

The results are summarized in Figures \ref{fig:speed}-\ref{fig:memory} and Table \ref{tab:memory}. All experiments were run on a single 8-core Intel Xeon 2.20 GHz CPU.

\begin{table}[htbp]
\centering
    \resizebox{\linewidth}{!}{
    \begin{tabular}[width=\textwidth]{lccccccc}
        \toprule
            &   $1\texttt{-D}_G$    &   $1\texttt{-D}_T$    &   \texttt{MVC}    &   \texttt{Gumbel} &     \texttt{Vickrey}    &   \texttt{TEM}    &   \texttt{Mahalanobis}    \\
        \midrule
        total   &   0.05   &   0.01  &   206.89   &   118.96   &  170.44   &   81.34 &   78.08 \\
        per-word   &  0.00005  &   0.00001 &   0.207  &   0.119  &   0.170   &  0.081  &   0.078   \\
       \bottomrule
    \end{tabular}
    }
    \caption{Memory consumption (in MiB) for 1000 words perturbed. Note that for both \textsc{1-Diffractor} settings, the consumption does not include the initial list configuration, as this is a one-time cost, and does not accumulate per word.}
    \label{tab:memory}
    \vspace{-1em}
\end{table}

\section{Discussion}
\label{sec:discuss}

\paragraph{Analysis of Experiments}

Leading the discussion on the merits of \textsc{1-Diffractor}, analysis begins with the basis of our one-dimensional word lists at the basis of the mechanism. The effect can be seen in Figure \ref{fig:nwsw}, where including double the number of lists (\textbf{L1}) provides a slight increase in privacy (plausible deniability) over only one list per model (\textbf{L0}). This comes with a negligible difference in utility loss (Figure \ref{fig:utility_delta}), thus prompting further investigation into the usage of an even greater number of lists simultaneously. 

With the use of only \textit{1 model, 1 list} (\textbf{L2}) in the case of both $\textsc{1-D}_G$ and $\textsc{1-D}_T$, the \textbf{L2} configuration provides clearly better utility preservation, at the cost of lower privacy protection, as shown in Figure \ref{fig:nwsw}. Thus, one can begin to observe the notion that while a greater number of (embedding) lists may increase empirical privacy, a carefully selected single list may be best for utility preservation.

Looking to the impact of $\varepsilon$, the utility performance across $\varepsilon$ values exhibit a clear \say{privacy-utility trade-off curve}, with an average utility drop of around 15 percentage points at 0.1, and near baseline performance for values above 5. The benefit of \textsc{1-Diffractor} is made salient by these results: within a relatively small range of $\varepsilon$ values, one can observe high levels of perturbation (measured by $N_w$) and considerably lower utility scores, and vice versa. The effectiveness of this obfuscation is solidified by the empirical privacy results, showing that it reduces the predictive power of adversaries.

As opposed to other tested mechanisms, \textsc{1-Diffractor} exhibits a much \say{tighter} $\varepsilon$ region, made most clear by Figures \ref{fig:nwsw} and \ref{fig:ep}. This can be attributed to the reduction to one dimension, where the overall magnitude of noise added is less as compared to MLDP mechanisms operating on distance metrics in higher dimensions. The simplicity of \textsc{1-Diffractor} thus leads to much smaller, interpretable privacy budgets, which follow a \say{graceful} degradation in privacy (and increase in utility) as the $\varepsilon$ value increases. While one may argue that this is simply a matter of scale, the benefit of such a bounded region can be interpreted as a clearer range of acceptable privacy budgets. This would certainly be necessary for adoption into practice.

\paragraph{The Privacy-Efficiency Trade-off}
The design of MLDP mechanisms often is evaluated for utility and privacy, sparking debates on the privacy-utility trade-off, yet the literature has largely ignored the question of efficiency, which would reasonably be required for such mechanisms to be employed in practice and at scale.

Our results show that \textsc{1-Diffractor} greatly outperforms previous methods in terms of the speed at which words are perturbed. This is particularly the case with our $\textsc{1-D}_G$ variant, which achieves immense speedups over previous MLDP mechanisms: over 15x from a theoretical estimate and over 90x from an empirical measurement (see Figure \ref{fig:speed}). Furthermore, these speedups can be realized on everyday hardware, i.e., a standard laptop CPU.

Such a speedup of course comes with a trade-off. As shown in our privacy experiments, $\textsc{1-D}_G$ exhibits lower theoretical privacy guarantees (via plausible deniability statistics) and demonstrates less effectiveness at lowering adversarial advantage. This is placed in juxtaposition to $\textsc{1-D}_T$, which operates at a considerably slower rate, yet demonstrates higher privacy benchmarks.

In the memory consumption benchmarks, \textsc{1-Diffractor} also shows significant improvements over previous mechanisms, even when accounting for the memory required to initialize the word lists (see Figure \ref{fig:memory}). This can be attributed to the fact that post-initialization, \textsc{1-Diffractor} does not need to perform expensive nearest neighbor searches for each word, which must be done in all other compared mechanisms. The resulting difference can be clearly observed in the per-word rates of Table \ref{tab:memory}. As can be seen in Figure \ref{fig:memory}, after only 2000 words perturbed, \textsc{1-Diffractor} already begins to use less memory than all other mechanisms.

\paragraph{The Question of an Optimal Obfuscation Mechanism}
In analyzing \textsc{1-Diffractor}, it is clear that even at lower $\varepsilon$ values, not as many tokens will be perturbed away from the original token. For example, at $\varepsilon=1$, only around 60\% of tokens are perturbed, as per the $N_w$ statistic, which is still significantly lower than other mechanisms even at higher values such as 10 (e.g., Mahalanobis $N_w @ 10 = 0.92$).

The debate here becomes how to interpret a \say{good} obfuscation mechanism. If nearly 100\% tokens are perturbed, this may grant high plausible deniability, but the utility will be impacted significantly. This is especially true in the case of high-dimensional perturbations, where privatized words may be far in meaning from the original word due to the effect of noise across many dimensions.


\citet{mattern-etal-2022-limits} echo the need for balanced obfuscation, namely not only with a high perturbation rate, but also in the preservation of semantic meaning and relatedly, utility in downstream tasks and empirically demonstrable privacy. In this work, we add to this notion, arguing that efficiency (speed) is also a key factor.

With \textsc{1-Diffractor}, we demonstrate such a balance, as the mechanism successfully preserves privacy via word perturbations, measured via plausible deniability and empirical privacy, while still allowing for the utility to remain intact. This is due not only to \textit{how often} words are perturbed, but also \textit{how} words are perturbed.

\paragraph{Open Challenges}
Our evaluation of \textsc{1-Diffractor} verifies its utility- and privacy-preserving capabilities, as well as a significant speedup over previous methods, yet a discussion of its merits must be accompanied by an analysis of open questions.

A major limitation of word-level MLDP methods comes with the inability to preserve grammatically correct sentence structures, due to the lack of context in single-word perturbations. To a degree, \textsc{1-Diffractor} preserves sentence coherence in the way that the lists are built: similar words functionally will (ideally) be sorted near each other. Nevertheless, this does not always hold, particularly at smaller $\varepsilon$ values.
The strength of \textsc{1-Diffractor} over previous MLDP mechanisms, however, is clearly demonstrated in our results. As an added limitation due to the word-level nature, our method cannot construct obfuscated outputs of differing length from the original sentence, another issue highlighted by \citet{mattern-etal-2022-limits}.

In continuing the discussion of Section \ref{sec:epsilon}, it is important to keep in mind the limitation of interpreting $\varepsilon$ across different MLDP mechanisms. The variation of underlying metrics makes evaluation challenging, as there exists no standard way of evaluating cross-metric DP mechanisms. Nevertheless, we address this shortcoming by testing on a wide range of $\varepsilon$ values. In this, we show that \textsc{1-Diffractor} remains competitive across all tested values, albeit with a diminishing theoretical privacy guarantee at higher values (i.e., 25 and 50). A concrete improvement is shown with \textsc{1-Diffractor}, in both versions, against the directly comparable TEM \cite{carvalho2023tem}, where our method consistently achieves higher utility scores while maintaining similar privacy levels. As mentioned above, this of course comes with the added benefit of higher perturbation speeds regardless of $\varepsilon$ value. Comparing once again to the original TEM of \citet{carvalho2023tem}, one can observe in Figure \ref{fig:speed} the effects of the dimensionality reduction offered by our list structure, where our $\textsc{1-D}_T$ performs significantly faster than the original TEM.

A final point comes with the question of dataset and task dependence. In our tests, we focus on a variety of tasks, from sentence similarity to textual entailment, yet we do not study the effect of each specific task. For example, one can see from Table \ref{tab:utility} that the entailment tasks are most affected in terms of utility (see MNLI and QNLI). In addition, regression tasks (i.e., STSB) are also affected more severely as opposed to classification. Factors like these call for more in-depth analyses of MLDP mechanism design and evaluation.

\section{Related Work}
\label{sec:related}

The notion of \textit{word-level} Metric (Local) DP was introduced by \citet{fernandes2019generalised}. This inspired several follow-up works \cite{feyisetan2019leveraging, feyisetan_balle_2020, xu2020differentially, xu2021density, xu2021utilitarian, carvalho2023tem}, which investigate the usage of various noise mechanisms or metric spaces. These works rely on embedding perturbations, which are carried out with DP to achieve private embeddings \cite{feyisetan2, klymenko-etal-2022-differential}. 

At the same time, other earlier works diverged from the word-level MLDP notion, focusing instead on private model training \cite{10.1145/2976749.2978318}, differentially private word replacement selection \cite{yue, chen-etal-2023-customized} or privacy-preserving neural representations of text \cite{10.1145/3342220.3344925, lyu-etal-2020-differentially}. A critique of earlier methods, particularly at the word level, by \citet{mattern-etal-2022-limits} highlights several shortcomings, as noted previously in this work. Other works \cite{feyisetan2021research, klymenko-etal-2022-differential} echo some of these challenges.

In light of these limitations, further works investigate the integration of DP in more advanced NLP models, such as earlier works on encoder(-decoder) models \cite{bo-etal-2021-er, ponomareva-etal-2022-training}. Other works extend beyond the word level to the sentence and document level \cite{meehan-etal-2022-sentence}. In recent state-of-the-art approaches, DP is achieved in combination with the training and fine-tuning of language models \cite{shi-etal-2022-selective, 10.1145/3576915.3616592}, or in directly adding noise to the latent representation, such as in DP-BART \cite{igamberdiev-habernal-2023-dp}.

Recent methods address the issue of DP-rewritten sentences with grammatical correctness \cite{ponomareva-etal-2022-training, wang2023differentially, yue2023synthetic, utpala-etal-2023-locally}. However, such methods rely on the utilization of computationally expensive language models, thus lacking scalability. Other methods, such as DP-BART \cite{igamberdiev-habernal-2023-dp}, rely on noise addition in high dimensions, thus leading to very large privacy budgets. Here, the \say{individual} more complex, as opposed to a word vocabulary with discrete and finite members.

Building upon these previous works, we follow in the footsteps of existing word-level MLDP mechanisms, focusing on efficiency while avoiding the usage of computationally expensive language models. In this way, we hope to advance the field of text privatization by emphasizing the design of utility- and privacy-preserving mechanisms that are lightweight and accessible to run.

\section{Conclusion}
\label{sec:conclusion}
In this work, we introduce \textsc{1-Diffractor}, a novel word-level MLDP mechanism for text obfuscation. \textsc{1-Diffractor} is built upon a simple and intuitive method of sorting words in one-dimensional lists, which serve as the basis for word privatization via Metric DP, achieved through a \textit{diffraction} of noise along this dimension. In a three-part evaluation, our method exhibits utility- and privacy-preserving capabilities, while notably demonstrating significant efficiency improvements over previous MLDP mechanisms. 

Our findings illustrate the merit of researching novel ways of representation for text privatization, showcasing that word-level perturbations are effective on the utility and privacy fronts, while also possessing the ability to be deployed at scale. This lightweightness makes a salient case for further research and future improvements.

We see three paths of further research to build upon our work, as well as its perceived limitations: (1) exploration into the effect of different word embedding models, as well as their combination and the use of multiple lists, (2) work on the creation of a uniform benchmark for word-level MLDP mechanisms, regardless of the underlying metric, and (3) relatedly, in-depth research into the design and implementation of utility \textbf{and} privacy-preserving mechanisms, that also can be practically deployed at scale.


\bibliographystyle{ACM-Reference-Format}
\bibliography{sample-base}

\appendix
\onecolumn
\section{Perturbation Examples}
\label{sec:appendix_ex}
Tables \ref{tab:examples1} and \ref{tab:examples2} show examples of perturbed sentences, taken from the validation set of \textsc{SST2}. Table \ref{tab:examples1} uses examples across different \textsc{1-Diffractor} configurations, while Table \ref{tab:examples2} compares \textsc{1-Diffractor} with the selected MLDP mechanisms, for chosen $\varepsilon$ values.

\begin{table*}[htbp]
    \centering
    \resizebox{0.8\linewidth}{!}{
\begin{tabular}{lll|l}
\toprule
\multicolumn{3}{r|}{Original sentence} & a delectable and intriguing thriller filled with surprises , read my lips is an original . \\
\multicolumn{3}{r|}{$\varepsilon$} &  \\ \hline
\multicolumn{1}{l|}{\multirow{9}{*}{$\textsc{1-D}_G$}} & \multicolumn{1}{c|}{\multirow{3}{*}{L0}} & 0.1 & a pet and impressed revisited powers with intrigues , reading my sight is an matched . \\
\multicolumn{1}{l|}{} & \multicolumn{1}{c|}{} & 1 & a delectable and fascinating remake fill with surprises , read my lips is an original . \\
\multicolumn{1}{l|}{} & \multicolumn{1}{c|}{} & 3 & a delectable and intriguing thriller filled with surprises , reading my lips is an original . \\
\cline{2-4} 
\multicolumn{1}{l|}{} & \multicolumn{1}{c|}{\multirow{3}{*}{L1}} & 0.1 & a buries and infatuation idling riddled with lend , reconsidered my stare is an tonnage . \\
\multicolumn{1}{l|}{} & \multicolumn{1}{c|}{} & 1 & a decadent and persuasive whodunit fill with surprises , write my cheeks is an new . \\
\multicolumn{1}{l|}{} & \multicolumn{1}{c|}{} & 3 & a delectable and intriguing thriller filled with surprises , read my lips is an original . \\
\cline{2-4}
\multicolumn{1}{l|}{} & \multicolumn{1}{c|}{\multirow{3}{*}{L2}} & 0.1 & a yikes and casinos whodunit excuses with unannounced , read my lipped is an territories . \\
\multicolumn{1}{l|}{} & \multicolumn{1}{c|}{} & 1 & a delectable and intriguing thriller riddled with surprises , read my lipped is an originals . \\
\multicolumn{1}{l|}{} & \multicolumn{1}{c|}{} & 3 & a delectable and intriguing thriller filled with surprises , read my lips is an original . \\
\hline \hline
\multicolumn{1}{l|}{\multirow{9}{*}{$\textsc{1-D}_T$}} & \multicolumn{1}{l|}{\multirow{3}{*}{L0}} & 0.1 & a heels and wooed surly adds with rooms , summarize my reveal is an mousy . \\
\multicolumn{1}{l|}{} & \multicolumn{1}{l|}{} & 1 & a sumptuous and intriguing suspense flooding with heartbreak , read my lips is an reactions . \\
\multicolumn{1}{l|}{} & \multicolumn{1}{l|}{} & 3 & a delectable and compelling thriller filled with surprises , read my lips is an original . \\
\cline{2-4} 
\multicolumn{1}{l|}{} & \multicolumn{1}{l|}{\multirow{3}{*}{L1}} & 0.1 & a butte and pleasant ebook interrogations with easter , xii my tooth is an willows . \\
\multicolumn{1}{l|}{} & \multicolumn{1}{l|}{} & 1 & a sumptuous and astounding blockbuster livestock with surprises , read my tassels is an original . \\
\multicolumn{1}{l|}{} & \multicolumn{1}{l|}{} & 3 & a sumptuous and intriguing suspense filled with surprises , read my cheek is an original . \\
\cline{2-4} 
\multicolumn{1}{l|}{} & \multicolumn{1}{l|}{\multirow{3}{*}{L2}} & 0.1 & a impossible and aroused spill torpedoed with huffy , reread my eyeliner is an mc . \\
\multicolumn{1}{l|}{} & \multicolumn{1}{l|}{} & 1 & a delicious and absorbing sherlock gb with boggles , reading my cheeks is an version . \\
\multicolumn{1}{l|}{} & \multicolumn{1}{l|}{} & 3 & a tasty and intriguing thriller filled with surprises , read my lips is an original . \\
\bottomrule
\end{tabular}
}
\caption{Examples of \textsc{1-Diffractor} perturbed sentences from the \textsc{SST2} validation set.} 
\label{tab:examples1}
\vspace{-2em}
\end{table*}

\begin{table*}[htbp]
    \centering
    \resizebox{0.89\linewidth}{!}{
\begin{tabular}{lll|l}
\toprule
\multicolumn{3}{r|}{Original sentence} & this one is definitely one to skip , even for horror movie fanatics . \\
\multicolumn{3}{r|}{$\varepsilon$} &  \\ \hline
\multicolumn{1}{c|}{\multirow{3}{*}{$\textsc{1-D}_G$ (L0)}} & \multicolumn{2}{c|}{0.1} & this forgot is went one to primary , barely for dread diana crowd . \\
\multicolumn{1}{l|}{} & \multicolumn{2}{c|}{1} & this these is surely one to skipped , even for horror movies fanatic . \\
\multicolumn{1}{l|}{} & \multicolumn{2}{c|}{10} & this one is definitely one to skip , even for horror movie fanatics . \\ 
\multicolumn{1}{l|}{} & \multicolumn{2}{c|}{50} & this one is definitely one to skip , even for horror movie fanatics . \\ \cline{2-4} 
\multicolumn{1}{c|}{\multirow{3}{*}{$\textsc{1-D}_T$ (L0)}} & \multicolumn{2}{c|}{0.1} & this shortly is indeed adamant to forsaking , shoved for tentacles adherents mistresses . \\
\multicolumn{1}{l|}{} & \multicolumn{2}{c|}{1} & this one is perhaps u to mana , even for movie movie fanatic . \\
\multicolumn{1}{l|}{} & \multicolumn{2}{c|}{10} & this one is definitely one to skip , even for horror movie fanatics . \\ 
\multicolumn{1}{l|}{} & \multicolumn{2}{c|}{50} & this one is definitely one to skip , even for horror movie fanatics . \\ 
\hline \hline
\multicolumn{1}{c|}{\multirow{3}{*}{MVC}} & \multicolumn{2}{c|}{0.1} & babes psychotherapy relapsing tofu inning assam chanting , connolly kerala female tusk kilkenny . \\
\multicolumn{1}{l|}{} & \multicolumn{2}{c|}{1} & [REDACTED] thistle amateur pleasuring oakland 357 albion , curlers yang parrots photographer 1844 . \\
\multicolumn{1}{l|}{} & \multicolumn{2}{c|}{10} & an deduction perjury unpredictable biological pattern managerial , feels [REDACTED] powder kurdish alerted . \\ 
\multicolumn{1}{l|}{} & \multicolumn{2}{c|}{50} & this one is definitely one to skip , even available horrors movie fanatics . \\ \hline \hline
\multicolumn{1}{c|}{\multirow{3}{*}{Gumbel}} & \multicolumn{2}{c|}{0.1} & this same is obviously only up skipped , perhaps with horrifying movie fanatics . \\
\multicolumn{1}{l|}{} & \multicolumn{2}{c|}{1} & one another has obviously only make go , actually well horror films geeks . \\
\multicolumn{1}{l|}{} & \multicolumn{2}{c|}{10} & that one which always another able skip , even with horror flick fanaticism . \\
\multicolumn{1}{l|}{} & \multicolumn{2}{c|}{50} & yet two seems obviously only to skip , even making gory movie fanatics . \\ \hline \hline
\multicolumn{1}{c|}{\multirow{3}{*}{Vickrey}} & \multicolumn{2}{c|}{0.1} & synagogue monterey goodnight 747 chemotherapy hippest townhouse , elected websites methadone parasite mapping . \\
\multicolumn{1}{l|}{} & \multicolumn{2}{c|}{1} & tequila coeds mutual interceptions 180 curves huskies , counterattack autonomy battery mesmerizing amor . \\
\multicolumn{1}{l|}{} & \multicolumn{2}{c|}{10} & too nightcap clinically godmother equations new goats , breed motels serie [REDACTED] jammu . \\
\multicolumn{1}{l|}{} & \multicolumn{2}{c|}{50} & one only which certainly part let instance , though to ghost films maniacs . \\ \hline \hline
\multicolumn{1}{c|}{\multirow{3}{*}{TEM}} & \multicolumn{2}{c|}{0.1} & progressed paint nibble cuddly lyrics famished bitty , buffy feverish complaining mundo disciplined . \\
\multicolumn{1}{c|}{} & \multicolumn{2}{c|}{1} & journals hounded weaving nightfall direct pereira translator , cries rc reason turmoil streetwalker . \\
\multicolumn{1}{l|}{} & \multicolumn{2}{c|}{10} &this one is definitely one to skip , even for horror movie fanatics . \\
\multicolumn{1}{l|}{} & \multicolumn{2}{c|}{50} & this one is definitely one to skip , even for horror movie fanatics . \\ \hline \hline
\multicolumn{1}{c|}{\multirow{3}{*}{Mahalanobis}} & \multicolumn{2}{c|}{0.1} & dunkirk motels knobs kant negligent pound harmonies , snooty housing tularemia bose hardcore . \\
\multicolumn{1}{l|}{} & \multicolumn{2}{c|}{1} & integer deluded 2d cozying beast spill upholstery , ghz hereford 24 polynomial rebelling . \\
\multicolumn{1}{l|}{} & \multicolumn{2}{c|}{10} & whoopee leader celibate cobb building spent subtitle , midi immoral plant director ljubljana . \\
\multicolumn{1}{l|}{} & \multicolumn{2}{c|}{50} & this one is definitely one to skip , even selling zombie movie fanatics . \\
\bottomrule
\end{tabular}
}
\caption{Perturbation examples of \textsc{1-Diffractor} and other MLDP mechanisms from one sentence in the \textsc{SST2} validation set.} 
\label{tab:examples2}
\end{table*}

\end{document}